\documentclass{article}

\usepackage{arxiv}

\usepackage{float}
\usepackage{cite}
\usepackage{amsmath,amssymb,amsfonts}
\usepackage{graphicx}
\usepackage{textcomp}
\usepackage{color}
\usepackage{url} 
\usepackage{bm}
\usepackage{multirow}
\usepackage{booktabs}
\usepackage[caption=false,font=footnotesize]{subfig}
\usepackage[ruled,vlined]{algorithm2e}
\usepackage{algpseudocode} 
\usepackage{array}
\usepackage[hidelinks]{hyperref}
\usepackage[capitalize,nameinlink]{cleveref}
\usepackage{makecell}
\usepackage{booktabs}

\newcommand{\nothresholdfirst}{No-Threshold (NT) }
\newcommand{\nothreshold}{NT }
\newcommand{\nothresholdp}{NT}

\title{Improving Mental Health Screening and Early Risk Detection in Spanish}

\author{
    \href{https://orcid.org/0009-0003-6000-3828}{
        \includegraphics[scale=0.06]{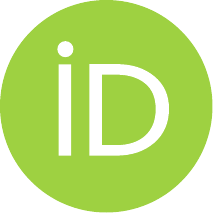}\hspace{1mm}
        Andreu Casamayor-Segarra
    } \\
    VRAIN: Valencian Research Institute for Artificial Intelligence \\
    Universitat Politècnica de València \\
    València, Spain \\
    \texttt{ancase3@upv.es}\\
    \AND
    \href{https://orcid.org/0000-0001-5636-651X}{
        \includegraphics[scale=0.06]{orcid.pdf}\hspace{1mm}
        Vicent Ahuir
    } \\
    VRAIN: Valencian Research Institute for Artificial Intelligence \\
    Universitat Politècnica de València \\
    València, Spain \\
    \texttt{vahuir@upv.es}
    \AND
    \href{https://orcid.org/0000-0001-6537-8803}{
        \includegraphics[scale=0.06]{orcid.pdf}\hspace{1mm}
        Antonio Molina-Marco
    } \\
    VRAIN: Valencian Research Institute for Artificial Intelligence \\
    Universitat Politècnica de València \\
    València, Spain \\
    \texttt{amolina@upv.es}
    \AND
    \href{https://orcid.org/0000-0002-1877-0455}{
        \includegraphics[scale=0.06]{orcid.pdf}\hspace{1mm}
        Lluís-F. Hurtado
    } \\
    VRAIN: Valencian Research Institute for Artificial Intelligence \\
    Universitat Politècnica de València \\
    ValgrAI: Valencian Graduate School and Research Network of Artificial Intelligence \\
    València, Spain \\
    \texttt{lhurtado@upv.es}
}

\renewcommand{\headeright}{}
\renewcommand{\shorttitle}{Improving Mental Health Screening and Early Risk Detection in Spanish}

\hypersetup{
    pdftitle={Improving Mental Health Screening and Early Risk Detection in Spanish},
    pdfsubject={cs.CL; cs.AI},
    pdfauthor={Andreu Casamayor-Segarra, Vicent Ahuir, Antonio Molina-Marco, Lluís-F. Hurtado},
    pdfkeywords={dataset relabeling, early detection, incremental context expansion, mental health, natural language processing, social media analysis, Spanish language models, transformers}
}
\begin{document}
\maketitle

\begin{abstract}
	Early detection of mental health disorders is often limited by the lack of specialized resources in Spanish and the difficulty of analyzing long histories of social media posts. This paper addresses these challenges through three main contributions. First, we introduce three Spanish foundational models specifically adapted to the mental health domain through domain-specific pre-training. Second, we propose Incremental Context Expansion (ICE), an automatic relabeling methodology designed for early detection. ICE identifies the point at which cumulative messages provide enough evidence of a disorder, generating more informative training samples. Third, we provide a set of fine-tuned models using the samples generated with the ICE methodology for early risk detection tasks. Our results on three Spanish benchmarks show that combining these specialized models with ICE improves the state-of-the-art, reducing detection latency while maintaining high performance. All models are publicly available.
\end{abstract}

\keywords{
Dataset relabeling
\and Early detection
\and Incremental context expansion (ICE)
\and Mental health
\and Natural language processing
\and Social media analysis
\and Spanish language models
\and Transformers
}

\section{Introduction}
Mental health disorders, including depression, anxiety, and schizophrenia, have become global issues affecting millions of lives. According to the World Health Organization (WHO), mental disorders comprise a group of clinically significant disturbances in cognition, emotional regulation, or behavior of an individual \cite{who2022}. Mental illness affects nearly one in eight people worldwide, but a large proportion of those affected are not diagnosed and consequently untreated. The prevalence of anxiety and depression increased by 26\% from 2019 to 2020. By 2021, an estimated 13.9\% of the global population was living with a mental health condition \cite{who2023mentalstats}. Furthermore, according to the European Commission’s 2023 report on mental health, 46\% of Europeans experienced an emotional or psychosocial problem, such as feeling depressed or anxious, during 2023 \cite{eurobarometer2023mentalhealth}. In Spain, according to the 2023 National Health System Report, approximately 34\% of the Spanish population suffers from a mental health disorder, surpassing 40\% in individuals over the age of 50, and more than 50\% in those aged 85 and above. Anxiety disorders are the most prevalent, affecting 10\% of the general population (14\% of women and 7\% of men) \cite{ministerio2023saludmental}.

The consequences of suffering from a mental illness can be severe, ranging from social isolation and a diminished quality of life to, in extreme cases, self-harm or suicide. Unfortunately, the stigma surrounding mental health often acts as a significant barrier, discouraging individuals from seeking help at an early stage. This highlights the pressing need for innovative techniques and technologies to facilitate early detection and intervention. 

Early identification of mental health issues is crucial for improving patient outcomes and alleviating broader social impacts. Conventional diagnosis methods, which rely on self-reporting and clinical assessment, may occasionally be delayed, reducing their potential implications for early intervention efforts. In contrast, analyzing publicly available social media data from platforms such as X and Reddit may facilitate the identification of early indicators of mental health distress. Social media are spaces where people openly share their emotions, thoughts, and personal problems, creating a rich source of real-time data \cite{coppersmith2015}. This data can be used to identify early indicators of mental illnesses and to foresee crises \cite{dechoudhury2013}. Researchers have already begun using social media posts to analyze trends in mental health \cite{ji2018supervised}. 

Over the last decade, transformer-based models \cite{vaswani2017} have become the dominant approach in natural language processing (NLP) over traditional machine learning methods. Before their introduction, most mental health detection systems relied on classical models such as support vector machines (SVMs) or logistic regression, often combined with TF–IDF features, n-grams, or handcrafted lexicons. Recent studies \cite{thakkar2024artificial} and shared tasks in the field, such as CLEF eRisk \cite{erisk2023} and IberLEF MentalRisk \cite{MentalRiskES2023}, have demonstrated consistent improvements when transitioning from traditional pipelines to deep contextual models. In particular, models such as BERT \cite{devlin2019} and RoBERTa \cite{liu2019} demonstrated substantial gains in symptom detection and user-level classification due to their ability to learn contextual, semantic, and affective information from large-scale corpora. These advances have positioned transformer-based language models (LMs) as key components in current mental health NLP research.

One of the challenges associated with using social media data is the large volume of posts that a single user can generate. This highlights a limitation of these models, which have a restricted input length and, therefore, are unable to process extended context effectively. To address this issue, advanced LMs like Longformer have been proposed \cite{beltagy2020}. These models adapt and extend the standard transformer architecture to handle longer text sequences efficiently, thus offering a more suitable solution in scenarios involving large user-level contexts \cite{beltagy2020}. 

Research on mental health detection in Spanish remains limited due to the scarcity of domain-specific resources and language models in this field. Most existing approaches focus on English, and multilingual models such as mBERT \cite{devlin2019} and XLM-R \cite{conneau2020} often show lower performance than their monolingual counterparts. This gap is mainly related to cross-lingual transfer issues \cite{wu2020}, since multilingual models struggle to transfer knowledge to languages that are poorly represented during pre-training \cite{lauscher2020}. These limitations underscore the need to develop language-specific models tailored to Spanish and the characteristics of mental health discourse.

The objectives of this work are aligned with the two main challenges described above: (1) the lack of Spanish language models adapted to the mental health domain, and (2) the need for effective strategies to handle early detection tasks involving long and evolving user message histories. Our goal is therefore twofold: to develop Spanish foundational models specialized in mental health, and to propose a dataset relabeling methodology that improves early risk detection across different disorders.

The contributions of this work are the following:

\begin{itemize}
    \item \textbf{Three Spanish foundational models tailored for mental health tasks:} These models were evaluated across various mental health-related tasks, and their construction and training process are detailed in subsequent sections.
    \item \textbf{A dataset relabeling methodology for mental illness early detection tasks:} We present the Incremental Context Expansion (ICE), an automatic relabeling methodology designed for early detection tasks. This methodology aims to increase the labeling information for enhancing the detection model's abilities in two key areas: (1) identifying when initial symptoms first appear within a sequence of messages, and (2) determining whether additional context is needed to confirm the detected symptoms confidently.
    \item \textbf{A set of fine-tuned models adapted to mental illness early detection tasks:} These models were trained with datasets modified using the ICE methodology and were evaluated across well-known Spanish mental illness early detection tasks.
    
\end{itemize}

We have made all the developed models publicly available through Hugging Face's model repository\footnote{  \url{https://huggingface.co/collections/ELiRF/mental-health}}.

The rest of the article is structured as follows. In \cref{sec:related_work}, we present a selection of relevant studies related to our work and identify opportunities for improvement. \cref{sec:datasets_tasks} details all the datasets used in this work and the classification tasks that we approached. In \cref{sec:foundational_models}, we detail the development of Spanish foundational models tailored for mental health applications. \cref{sec:ice_method} details of our proposed relabeling methodology aimed at early detection tasks within the context of mental health. In \cref{sec:discussions}, we highlight the main aspects extracted from the results, as well as the strengths and limitations of our approach. We end with the conclusions (\cref{sec:conclusions}) and future work (\cref{sec:future_work}).

\section{Related Work}
\label{sec:related_work}

Research on NLP for mental health has grown significantly over the past years, supported by several surveys and reviews that summarize the main advances in this field \cite{thakkar2024artificial}. Within this broad literature, we focus on three representative works, selected because they address key aspects relevant to our approach: domain-adapted language models for mental health, resources for Spanish or multilingual settings, and early detection methods based on user timelines. Together, they provide the context needed to position our models and the strategy proposed in this paper.

The work by \textbf{S. Ji, T. Zhang, L. Ansari, J. Fu, P. Tiwari, and E. Cambria} \cite{DBLP:journals/corr/abs-2110-15621} has been a key source of inspiration for the design of our models. In this study, the authors introduce MentalBERT and MentalRoBERTa, two language models based on BERT and RoBERTa, respectively, which were pre-trained on a dataset of more than 13.6 million sentences collected from Subreddits focused on mental health, such as r/depression, r/Anxiety, and r/SuicideWatch. Using a continuous pre-training strategy, they were able to adapt general models to a domain as specific as mental disorders, achieving notable improvements over biomedical models such as BioBERT and ClinicalBERT. Adopting a similar approach, we have incorporated the methodological principles of MentalBERT into our work, both in the development of our foundational models and in preparing the translated dataset for their pre-training.

The study by \textbf{J. Martinez-Romo, L. Araujo, and B. Reneses} \cite{martinez2025guardian} presents a domain-adapted BERT-based language model developed for the early detection of self-injury and suicidal tendencies in Spanish clinical reports. The model follows a two-phase training strategy: an initial pre-training on general medical records, followed by fine-tuning on psychiatric-specific clinical data. Evaluated on real-world clinical notes from Hospital Clínico San Carlos (Madrid), Guardian-BERT demonstrated strong performance, achieving an F1-score of 0.95 for non-suicidal self-injury (NSSI) and 0.89 for suicidal behavior. In addition to its predictive capabilities, the model incorporates risk factor analysis, such as substance abuse, family conflict, and social vulnerability, thereby enhancing its interpretability and clinical utility.

The work by \textbf{M. Couto, A. Perez, J. Parapar, and D. E. Losada} \cite{couto2025temporal}, introduce a novel framework for the early detection of psychological disorders on social media by modeling how language evolves. The authors propose two distinct methodologies for learning temporal word representations: TWEC (Temporal Word Embeddings with a Compass), a Word2Vec-based model that maintains semantic alignment across discrete time slices through a compass vector, and DCWE (Dynamic Contextualized Word Embeddings), a BERT-based approach that dynamically adjusts contextualized word embeddings as user language shifts. Their experiments reveal that both methods are effective for early classification tasks, with DCWE consistently outperforming TWEC, particularly in terms of precision and timeliness.

Furthermore, the study incorporates a lexical-semantic analysis comparing symptomatic users and control groups, highlighting how specific keywords suffer semantic drift over time in users displaying mental health symptoms. This behavior supports the underlying hypothesis that linguistic changes can serve as early indicators of psychological decline. Due to its innovative approach and its focus on temporal language modeling, we have drawn on this work to inform our own strategy for tackling early detection of mental health disorders.

Despite these advancements, several limitations persist in the current landscape of mental health detection using NLP. As seen in two of the previously discussed articles, MentalBERT and TWEC, the proposed approaches are developed exclusively for English, leaving other languages, such as Spanish, largely overlooked. This widens the gap between high-resource and low-resource languages, reinforcing the need for further research and model development tailored to linguistic communities with limited resources compared to English.

Although Guardian-BERT stands as a valuable contribution in the Spanish context, it focuses solely on clinical environments, being trained and evaluated using formal clinical reports. This means it relies on pre-existing medical documentation for inference, limiting its applicability to broader, real-world scenarios. In contrast, our work presents models specifically designed to detect mental health conditions in Spanish-language social media. By targeting this domain, we address both the linguistic gap and the need for early detection methods that operate outside clinical settings.

In recent years, eRisk (Early Risk Prediction on the Internet) competitions have established themselves as a key benchmark in the field of NLP applied to the early detection of mental health risks. Organized within the framework of the CLEF (Conference and Labs of the Evaluation Forum) conference, these competitions focus on predicting conditions such as depression, anxiety disorders, and suicide risk based on the analysis of texts published on social media, mainly Reddit. A distinctive feature of eRisk is its evaluation structure based on early prediction: systems must issue a decision as user content is progressively revealed to them, simulating a realistic scenario of continuous monitoring \cite{erisk2017, erisk2018, erisk2019, erisk2020, erisk2021, erisk2022}.

At the same time, the MentalRisk competition, organized within the framework of IberLEF (Iberian Languages Evaluation Forum), presents similar challenges in the field of mental health, but with a focus on analyzing texts written in Spanish. MentalRisk sets tasks such as the early detection of various mental disorders, including depression, anxiety, self-harm, and eating disorders, based on posts taken from Telegram. Unlike eRisk, which has traditionally focused on tasks in English, MentalRisk encourages the development of models specifically adapted to the Spanish-speaking linguistic and cultural context \cite{MentalRiskES2023}.

In addition to serving as a reference in the state of the art, the eRisk and MentalRisk competitions have been fundamental to the development of our work. Based on the analysis of tasks, methodologies, and results reported in these editions, we have identified successful approaches and recurring challenges in the automatic detection of mental disorders. This has enabled us to develop solutions tailored to the frameworks proposed by both competitions. Likewise, the official datasets and metrics provided by eRisk and MentalRisk have been used to validate our proposals, ensuring a rigorous evaluation that is comparable with other systems presented in these initiatives.

\section{Datasets and Tasks}
\label{sec:datasets_tasks}

\label{tasks}

This work focuses on three tasks proposed in the 2023 and 2024 editions of the MentalRisk shared task \cite{marmol-romero-etal-2024-mentalriskes}, organized within the IberLEF conference \cite{IberLEF2024}. These tasks provide user-level datasets in Spanish, covering relevant problems in the mental health domain and placing special emphasis on early detection. They constitute an appropriate benchmark for evaluating models for mental health screening and early risk detection tasks in Spanish. 

\subsection{description of the tasks}

The MentalRisk shared tasks include subtasks targeting specific mental health conditions and following the early detection framework described above. Although they share a standard structure (user-level data, chronological posts, and binary or multiclass labels) each task presents its own characteristics and challenges. Below, we describe the three tasks considered in this study.

\subsubsection{MentalRisk24 - Disorder Detection}

\textit{``MentalRiskEs 2024 - Disorder Detection''} (MR24-DD) task focused on the detection of two clinically relevant mental disorders: depression and anxiety \cite{marmol2024overview}. The dataset included posts from Telegram written by multiple users, each labeled according to their condition: \textit{Depression}, \textit{Anxiety}, or \textit{None}, in cases where they did not present any of the mentioned diseases. Therefore, this task was approached as a multi-class classification problem.

\Cref{tab:est_DD} summarizes the descriptive statistics of the MR24-DD dataset. User activity exhibits a moderate level of variability, with users producing an average of 36 posts, most of which fall within the range of 18 to 42. Token counts are similarly dispersed, with a mean of 496 tokens per user. Messages remain concise across the dataset, averaging 13.7 tokens per post, indicating relatively consistent short-form communication patterns.

\begin{table}[h!]
\centering
\caption{Descriptive statistics for posts, tokens, and tokens per post across users in the MR24-DD.}
\begin{tabular}{lccc}
\toprule
\textbf{Statistic} & \textbf{Posts/user} & \textbf{Tokens/user} & \textbf{Tokens/post} \\ 
\midrule
Mean & 36.26 & 524.81 & 14.48\\
Standard deviation & 26.20 & 429.70 & 21.77 \\
Minimum & 10 & 74 & 1\\
25th percentile (P25) & 18 & 229 & 6 \\
Median (P50) & 29 & 396 & 9\\
75th percentile (P75) & 42 & 680 & 18 \\
Maximum & 100 & 3803 & 783 \\
\bottomrule
\end{tabular}
\label{tab:est_DD}
\end{table}

The MR24-DD original dataset was organized in three partitions: the Trial partition, which included 20 users; the Training partition, with 465 users; and the Test partition, consisting of 400 users. For our study, we merged the Trial and Train partitions into a single partition. Subsequently, we created two partitions in a stratified manner: 80\% of them for training (388 users) and 20\% for validation (97 users). \cref{Table:Dataset_MR24} presents the resulting partitions and their distribution by class.

\begin{table}[!ht]
\centering
\caption{Distribution of samples in the dataset for the MR24-DD task.}
\label{Table:Dataset_MR24}
\setlength{\tabcolsep}{10pt}
\renewcommand{\arraystretch}{1.3}
\begin{tabular}{lccc}
\hline
\textbf{Class} & \textbf{Training} & \textbf{Validation} & \textbf{Test} \\
\hline
None & 178 & 45 & 200 \\
Depression & 134 & 35 & 100 \\
Anxiety & 76 & 17 & 100 \\
\hline
\textbf{Total} & \textbf{388} & \textbf{97} & \textbf{400} \\
\hline
\end{tabular}
\end{table}

\subsubsection{MentalRisk23 - Eating Disorders}

The \textit{``MentalRisk23 - Eating Disorders''} (MR23-ED) task focused on the early detection of eating disorders, specifically anorexia and bulimia \cite{MentalRiskES2023}. The data was collected from public Telegram groups. The labeling was done at the user level, where each individual was classified based on their condition regarding anorexia and bulimia. The task was framed as a binary classification problem for each eating disorder.

\Cref{tab:est_ED} summarizes the main statistics of the MR23-ED dataset. As shown in \cref{tab:est_ED}, users produced an average of 31 posts and 472 tokens, with considerable variability in both measures. Messages remained generally concise, averaging 14.6 tokens per post, although a few users exhibited substantially higher verbosity.

\begin{table}[h!]
\caption{Descriptive statistics for posts, tokens, and tokens per post across users in the MR23-ED dataset}
\centering
\begin{tabular}{lccc}
\toprule
\textbf{Statistic} & \textbf{Posts/user} & \textbf{Tokens/user} & \textbf{Tokens/post} \\
\midrule
Mean & 31.34 & 472.46 & 14.65 \\
Standard deviation & 15.33 & 612.45 & 12.93 \\
Minimum & 11 & 76 & 6.00 \\
25th percentile (P25) & 17 & 207 & 9.64 \\
Median (P50) & 28 & 358 & 12.17 \\
75th percentile (P75) & 50 & 568.5 & 15.51 \\
Maximum & 50 & 7396 & 147.92 \\
\bottomrule
\end{tabular}
\label{tab:est_ED}
\end{table}

The MR23-ED original dataset consisted of three partitions: Trial (with 10 users), Train (with 175 users), and Test (with 150 users). We combined Trial and Train and split them in a stratified manner: 80\% for training (148 users) and 20\% for validation (37 users). \Cref{Table:Dataset_ED} presents the class distribution for the resulting partitions.

\begin{table}[!ht]
\centering
\caption{Distribution of samples in our dataset for the MR23-ED task.}
\label{Table:Dataset_ED}
\setlength{\tabcolsep}{12pt}
\renewcommand{\arraystretch}{1.3}
\begin{tabular}{lccc}
\hline
\textbf{Class} & \textbf{Training} & \textbf{Validation} & \textbf{Test} \\
\hline
None    & 85  & 22  & 86 \\ 
ED      & 63  & 15  & 64 \\ 
\hline
\textbf{Total} & \textbf{148} & \textbf{37} & \textbf{150} \\
\hline
\end{tabular}
\end{table}

\subsubsection{MentalRisk23 - Depression}

The \textit{``MentalRisk23 - Depression''} (MR23-D) task focused on the early detection of symptoms of depression in users considering their activity on the Telegram social network \cite{MentalRiskES2023}. Each sample of the dataset contained the messages posted for a user and whether that activity showed signs of depression or not.

\Cref{tab:est-D} presents the descriptive statistics of the MR23-D dataset. On average, users produced 36 posts and 503 tokens, although both measures display substantial dispersion across the population. Message length follows a similar pattern: while the median remains relatively concise at 12 tokens per post, values can reach up to 95 tokens, revealing a wide range of communication styles within the dataset.

\begin{table}[h!]
\caption{Descriptive statistics for posts, tokens, and tokens per post across users in the MR23-D dataset.}
\centering
\begin{tabular}{lccc}
\toprule
\textbf{Statistic} & \textbf{Posts/user} & \textbf{Tokens/user} & \textbf{Tokens/post} \\
\midrule
Mean & 36.02 & 503.45 & 15.06 \\
Standard deviation & 25.99 & 428.94 & 10.88 \\
Minimum & 11 & 75 & 4.80 \\
25th percentile (P25) & 19 & 228 & 9.16 \\
Median (P50) & 28 & 370 & 11.99 \\
75th percentile (P75) & 41 & 640 & 16.99 \\
Maximum & 100 & 3803 & 95.08 \\
\bottomrule
\end{tabular}
\label{tab:est-D}
\end{table}

The original dataset was divided into three partitions: Trial (with 10 users), Training (with 175 users), and Test (with 149 users). We combined Trial and Train and split them in a stratified manner: 80\% for training (148 users) and 20\% for validation (37 users).
\Cref{Table:Dataset_Depression} presents the class distribution for the resulting partitions.

\begin{table}[!ht]
\centering
\caption{Distribution of samples in our dataset for the MR23-D task.}
\label{Table:Dataset_Depression}
\setlength{\tabcolsep}{10pt}
\renewcommand{\arraystretch}{1.3}
\begin{tabular}{lccc}
\hline
\textbf{Class} & \textbf{Training} & \textbf{Validation} & \textbf{Test} \\
\hline
None       & 70  & 15  & 86 \\ 
Depression        & 78  & 22  & 64 \\ 
\hline
\textbf{Total} & \textbf{148} & \textbf{37} & \textbf{150} \\
\hline
\end{tabular}
\end{table}

Considering the statistics reported for the three tasks, it becomes evident that many users accumulate a substantial number of tokens throughout their message histories. Although individual posts tend to be short, the complete user context frequently surpasses the 512-token limit imposed by standard transformer architectures. In early-detection settings, where the chronological order of messages must be preserved, and information loss can negatively affect model performance, such limitations would lead to repeated truncation of the input. For this reason, we also incorporate Longformer-based models, as their extended context capacity enables the system to process longer user message histories and more accurately capture the temporal evolution of linguistic signals.

\subsection{Early Risk Detection Task Protocol}
\label{format_competi}

The three tasks described above follow the evaluation protocol established by the MentalRisk shared tasks. Unlike conventional classification settings, where the full user history is available at once, early detection is framed as a sequential decision-making problem. Following the official methodology published by the organizers\footnote{\url{https://github.com/sinai-uja/MentalRiskES-IberLEF/tree/main}}, the evaluation is structured into multiple rounds. In each round, the system receives one additional post from every user, presented in chronological order, and must decide, based only on the information observed so far, whether the user shows signs of a mental disorder.

At each step, the model may either predict the domain-specific risk label (e.g., at-risk, clinical condition) or choose the label 'none' if it lacks sufficient evidence to issue a positive decision. Importantly, once a user is predicted as positive at any round, that prediction becomes final and cannot be modified in later rounds. Conversely, users who are never classified as positive across the entire sequence are considered non-risk cases for evaluation purposes.

This protocol simulates a real-time monitoring scenario, where systems must balance accuracy with timeliness, issuing early and reliable predictions while avoiding premature or incorrect risk detection.

\subsection{Metrics}

Replicating the evaluation protocol used in the MentalRisk competitions, we adopt the following metrics to assess system performance:

\begin{itemize}
    \item \textbf{F1-score:} The Macro F1-score is used as the main classification metric. It computes the F1-score independently for each class and then averages the results, giving the same weight to all classes regardless of their frequency. This makes it suitable for datasets with class imbalance, as is often the case in the MentalRisk tasks.
    \item \textbf{ERDE:} The Early Risk Detection Error (ERDE) metric \cite{ERDE} is specifically designed for early-detection scenarios. It penalizes both incorrect predictions and correct predictions that arrive too late. The earlier and more accurate a model’s decision, the lower its ERDE value. In this work, we report two variants: \texttt{ERDE5} and \texttt{ERDE30}, which apply penalization after 5 and 30 time units or rounds, respectively. 

    \item \textbf{LTP:} \textit{Latency True Positive} (LTP) measures the average position (i.e., the round) at which the model issues its first correct positive prediction. Lower values indicate that the system can detect at-risk users earlier in their message sequence. This metric complements ERDE by providing a more direct interpretation of timeliness, independent of the error function’s penalization scheme.
\end{itemize}

\subsection{Reference Systems}

Establishing appropriate reference systems is essential to ensure a fair and interpretable evaluation of the proposed models. For each task, we consider a set of representative systems in terms of performance or architecture:

\begin{itemize}
    \item \textbf{MR24-DD:} We selected a system based on the Longformer architecture pre-trained in English \cite{Casamayor2024}. This model processes all user messages translated into English and applies a fixed minimum context of 100 words before issuing predictions. We also included the system proposed by the UNED-GELP team \cite{UNED-GELP}, which follows a two-step pipeline built upon the BETO model \cite{BETO}. Finally, we included the official RoBERTa-based baseline model released by the organizers, which incorporates a mechanism specifically designed to better capture early-detection dynamics.

    \item \textbf{MR23-ED:} We chose the best performing system of the task \--- the CIMAT-NLP-GTO system \cite{Echeverria-Baru2023} \---, which adopts a classical approach based on TF-IDF representations combined with a Naive Bayes classifier. As a transformer-based reference system, we included the RoBERTa-large baseline model provided by the competition organizers. We also considered the UNSL system \cite{UNSL}, which achieved the best performance among transformer-based systems according to the \textit{ERDE30} and \textit{F1-score} metrics.

    \item \textbf{MR23-D:}  We selected the system developed by the UMUTeam \cite{Pan2023UMUTeam}, which integrates transformer-based fine-tuning with ensemble learning techniques. We also included two additional transformer systems: the VICOM-nlp model \cite{Vicon}, based on a fine-tuned BERT encoder, and the SINAI-SELA system \cite{SINAI}, which relies on a fine-tuned BETO architecture.
\end{itemize}

\section{Spanish Foundational Models for Mental Health}
\label{sec:foundational_models}

\begin{figure}[H]
    \centering
    \caption{Workflow for adapting Spanish LMs to the mental health domain.}
    \includegraphics[width=0.6\linewidth]{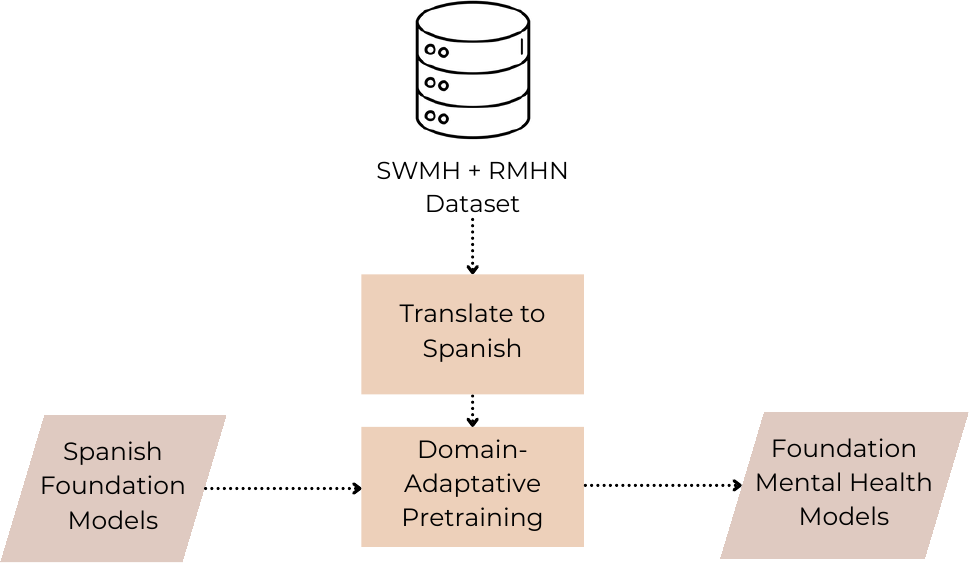}
    \label{fig:foundations-models}
\end{figure}

Developing effective systems for early detection of mental health conditions requires language models that are not only capable of processing Spanish text but also adapted to the specific linguistic characteristics of mental health discourse. General-purpose Spanish models, although strong in broad NLP tasks, do not fully capture these domain-specific patterns. For this reason, we developed a set of foundational models tailored to the mental health domain through domain-adaptive pre-training (DAP) \cite{DBLP:journals/corr/abs-2004-10964}. The general workflow followed in this process is illustrated in \cref{fig:foundations-models}.

The \cref{fig:foundations-models} outlines the process for adapting Spanish LMs to the mental health domain. Starting from a dataset of mental health–related posts translated into Spanish, we applied DAP to extend the capabilities of existing Spanish language models. The resulting models were later evaluated on several downstream mental health tasks to determine their suitability for both standard classification and early detection scenarios.

\subsection{Domain-Adaptive Pretraining (DAP)}
\label{Pre-training-phase}
To build our mental health foundational models, we employed DAP using a set of existing language models and a compilation of mental health-related data for domain adaptation. DAP aimed to align general language models with the linguistic and conceptual nuances of the mental-health discourse.

\subsubsection{Base Language Models}

To create solid metal-health-related foundational models, we selected the MarIA Spanish models \cite{MarIA} due to their extensive, diverse training data, multiple architectures (RoBERTa and Longformer), and strong benchmark results on a range of NLP tasks. The selected base language models were:

\begin{itemize}
    \item \textbf{PlanTL-GOB-ES/roberta-large-bne} {\cite{MarIA}}: This model is a transformer-based masked language model specifically designed for the Spanish language. Developed on the architecture of the RoBERTa large model, it has been pre-trained on Spanish corpora extracted from the National Library of Spain. This model has 355M parameters.

    \item \textbf{PlanTL-GOB-ES/Longformer-base-4096-bne-es}\footnote{https://huggingface.co/PlanTL-GOB-ES/longformer-base-4096-bne-es}: Is an alternate adaptation of the Spanish-language-specific roberta-base-bne model, employing Longformer architecture. This model is capable of processing much larger input contexts. Initializations were performed using the Roberta-base-bne checkpoint; pre-training employed a masked language modeling objective on long-form documents from the National Library of Spain. The model has 149M parameters. 
\end{itemize}

\subsubsection{Dataset}

To adapt these base LMs to the mental health domain, we chose two existing datasets: Reddit SuicideWatch and Mental Health Collection (SWMH), and Reddit Mental Health Narratives (RMHN). The SWMH dataset \cite{dataset} was explicitly designed to train language models for various mental health conditions. It contains writings from emotional support forums and transcripts of therapeutic interventions. The dataset consists of text snippets extracted from different Subreddits, including communities such as 'r/depression' or 'r/anxiety'.  The RMHN dataset \cite{app14041547} was designed to capture linguistic patterns associated with mental health conditions during the COVID-19 pandemic. It contains user-generated posts from multiple emotional support communities such as 'r/lonely', 'r/bipolar', and 'r/mentalhealth', annotated with thematic and emotional labels. The dataset provides a diverse and realistic representation of online mental health discourse, making it suitable for domain-adaptive pre-training of language models.

All texts were automatically translated into Spanish using EasyNMT \cite{bulgarelli2021easynmt} with the OPUS-MT models \cite{tiedemann2023democratizing, TiedemannThottingal:EAMT2020}, which provide robust neural machine translation for multiple language pairs. After translation, we merged both datasets to obtain a unified dataset for domain-adaptive pre-training.

\Cref{tab:combined_label_distribution} summarizes the distribution of posts across the seven mental health communities included in the combined SWMH + RMHN dataset. Most posts come from "r/depression" and "r/SuicideWatch", which together represent nearly 60\% of the dataset. Communities such as "r/mentalhealth", "r/Anxiety", and "r/lonely" provide additional coverage of diverse user experiences, while "r/offmychest" and "r/bipolar" appear less frequently. In total, the dataset comprises around 1.9M posts, offering broad and representative coverage of mental health discourse for domain-adaptive pre-training

\begin{table}[h!]
\centering
\caption{Distribution of posts across subreddits in the combined SWMH + RMHN dataset.}
\label{tab:combined_label_distribution}
\begin{tabular}{lrr}
\toprule
\textbf{Subreddit} & \textbf{Samples} & \textbf{Percentage} \\
\midrule
depression     & 643,307 & 33.81\% \\
SuicideWatch   & 493,230 & 25.94\% \\
Anxiety        & 289,593 & 15.22\% \\
mentalhealth   & 303,109 & 15.93\% \\
lonely         & 157,114 &  8.26\% \\
offmychest     &   8,284 &  0.44\% \\
bipolar        &   7,645 &  0.40\% \\
\midrule
\textbf{Total} & \textbf{1,902,282} & \textbf{100\%} \\
\bottomrule
\end{tabular}
\end{table}

\Cref{tab:stats_combined} summarizes the statistics of the combined dataset. The average post length is 154 tokens, with a median of 97, and the distribution exhibits a long tail, reaching up to 9,680 tokens. The 75th and 90th percentiles (198 and 352 tokens) indicate that many posts exceed the typical 128–512 token limits of standard transformer models, motivating the use of long-context architectures such as Longformer.

\begin{table}[h!]
\centering
\caption{Token statistics per post of the combined SWMH + RMHN dataset.}
\label{tab:combined_length_stats}
\begin{tabular}{lr}
\toprule
\textbf{Statistic} & \textbf{Tokens} \\
\midrule
Mean & 154.48 \\
Standard deviation & 207.63 \\
Median & 97 \\
75th percentile  & 198 \\
90th percentile  & 352 \\
Maximum & 9,680 \\
\bottomrule
\end{tabular}
\label{tab:stats_combined}
\end{table}

\subsubsection{Resulting Foundational Language Models}
\label{sec:foundational_models_m}

A key contribution of this work is the development of three Spanish domain-specific foundational models for mental health. These models were obtained through DAP using the translated datasets mentioned above. The models were continued pre-trained using 4 Nvidia A40 GPUs for 20 epochs with a batch size of 8 per GPU, and evaluations were performed every 10,000 steps. The entire pre-training process took approximately four days to complete. The resulting foundational models for mental health are the following:

\begin{enumerate}
    \item \textbf{RoBERTa-es-mental-large:} RoBERTa-like model for mental health with a maximum of 512 input tokens. The model was the result of applying DAP to the RoBERTa-large-bne general language model.
    
    \item \textbf{Longformer-es-mental-base:} Longformer-like model for mental health with a maximum of 4096 input tokens. This model was obtained by applying DAP to the  Longformer-base-4096-bne general language model.
    
    \item \textbf{Longformer-es-mental-large:} Longformer-like model for mental health with a maximum of 4096 input tokens. This model was obtained by adapting the RoBERTa-es-mental-large to the Longformer architecture. 
\end{enumerate}

\cref{Table:resume_models} summarizes the architecture type, the number of maximum input tokens accepted by the model, and the size of the model.

\begin{table}[!ht]
\caption{Foundational Models for Mental Health in Spanish}
\label{Table:resume_models}
\centering
\renewcommand{\arraystretch}{1.3}
\setlength{\tabcolsep}{2pt}
\begin{tabular}{lcccc}
\hline
\textbf{Models} & \textbf{Arquitecture} & \textbf{Tokens} & \textbf{Parameters} \\
\hline
{RoBERTa-es-mental-large} & RoBERTa & 512 & 355M \\
{Longformer-es-mental-base} & Longformer & 4096 & 149M \\
{Longformer-es-mental-large} & Longformer & 4096 & 435M \\
\hline
\end{tabular}
\end{table}

\subsection{ Evaluation on Standard User-Level Classification}

The evaluation aims to assess how effectively the proposed models capture mental health-related signals compared to their reference systems counterparts. We conducted user-level classification experiments, a straightforward approach that considers the complete user context when making predictions, using well-established tasks and metrics to ensure comparability.

\subsubsection{Models for Downstream Tasks}
\label{sec:models_ft_1}

To obtain the classification models for the downstream tasks, we added a feed-forward layer as a classification layer that receives the embedding from the last attention layer of the \texttt{[CLS]} token to classify. The task-specific models were obtained by fine-tuning them without frozen layers. As input, these models received the full available user context for each instance, resulting in predictions made at the user level. All models were fine-tuned under this same configuration, while the best-performing epoch was selected individually for each case based on validation performance. \Cref{Table:Parametres_fine_tuning_1} presents the hyperparameters selected for the fine-tuning process.

\begin{table}[!ht]
\caption{Hyperparameters for the fine-tuning process.}
\label{Table:Parametres_fine_tuning_1}
\centering
\renewcommand{\arraystretch}{1.3}
\setlength{\tabcolsep}{10pt}
\begin{tabular}{p{4cm}r}
\hline
\textbf{Parameter} & \textbf{Value} \\
\hline
Optimizer               & AdamW    \\
Learning rate           & 5e-5     \\
LR scheduler type       & Linear   \\
Weight decay            & 0.01     \\
Number of epochs        & 10       \\
Training batch size     & 16       \\
\hline
\end{tabular}

\end{table}

\subsubsection{Metrics}

For the classification tasks, we evaluated the performance of the models using the following metrics: Precision, Recall, and F1-score.

\subsubsection{Results}

In this subsection, we present the results obtained by the models introduced in \Cref{Pre-training-phase} across the three evaluated tasks.

\begin{table}[!ht]
\centering
\caption{\textbf{Performance results of the fine-tuned models across the Test partition of the tasks.}}
\label{tab:results_tasks}
\setlength{\tabcolsep}{2.5mm}
\renewcommand{\arraystretch}{1.25}
\small

\begin{minipage}[t]{0.48\textwidth}
    \centering
    \begin{tabular}{lccc}
        \toprule
        \multicolumn{4}{c}{\textbf{MentalRisk24 - Disorder Detection}} \\
        \multicolumn{4}{c}{\textbf{(MR24-DD)}} \\
        \midrule
        \textbf{Model} & \textbf{Precision} & \textbf{Recall} & \textbf{F1} \\
        \midrule
        RoBERTa-large-bne       & 0.821 & 0.781 & 0.813 \\
        Longformer-base-bne     & 0.828 & 0.865 & 0.843 \\
        \midrule
        RoBERTa-es-m-large      & 0.854 & 0.849 & 0.852 \\
        Longformer-es-m-base    & 0.897 & 0.900 & 0.897 \\
        Longformer-es-m-large   & \textbf{0.901} & \textbf{0.909} & \textbf{0.907} \\
        \bottomrule
    \end{tabular}
\end{minipage}
\hfill
\begin{minipage}[t]{0.48\textwidth}
    \centering
    \begin{tabular}{lccc}
        \toprule
        \multicolumn{4}{c}{\textbf{MentalRisk23 - Eating Disorders}} \\
        \multicolumn{4}{c}{\textbf{(MR23-ED)}} \\
        \midrule
        \textbf{Model} & \textbf{Precision} & \textbf{Recall} & \textbf{F1} \\
        \midrule
        RoBERTa-large-bne       & 0.814 & 0.823 & 0.818 \\
        Longformer-base-bne     & 0.843 & 0.857 & 0.838 \\
        \midrule
        RoBERTa-es-m-large      & 0.871 & 0.876 & 0.873 \\
        Longformer-es-m-base    & 0.881 & 0.884 & 0.883 \\
        Longformer-es-m-large   & \textbf{0.918} & \textbf{0.925} & \textbf{0.920} \\
        \bottomrule
    \end{tabular}
\end{minipage}

\vspace{4mm}

\begin{minipage}[t]{0.48\textwidth}
    \centering
    \begin{tabular}{lccc}
        \toprule
        \multicolumn{4}{c}{\textbf{MentalRisk23 - Depression}} \\
        \multicolumn{4}{c}{\textbf{(MR23-D)}} \\
        \midrule
        \textbf{Model} & \textbf{Precision} & \textbf{Recall} & \textbf{F1} \\
        \midrule
        RoBERTa-large-bne       & 0.757 & 0.701 & 0.689 \\
        Longformer-base-bne     & 0.747 & 0.672 & 0.635 \\
        \midrule
        RoBERTa-es-m-large      & 0.768 & 0.730 & 0.746 \\
        Longformer-es-m-base    & 0.768 & 0.743 & 0.753 \\
        Longformer-es-m-large   & \textbf{0.786} & \textbf{0.756} & \textbf{0.767} \\
        \bottomrule
    \end{tabular}
\end{minipage}

\end{table}

\Cref{tab:results_tasks} summarizes the comparative performance of all domain-specific and general-domain models across the three evaluation tasks.

Across the three tasks, the mental-health-oriented foundational models consistently outperform their general-domain counterparts. This confirms the effectiveness of domain-adaptive pre-training for capturing the linguistic and semantic properties characteristic of mental-health discourse in Spanish. Among the proposed models, Longformer-es-m-large obtains the best overall results, reaching Macro-F1 scores of 0.907, 0.920, and 0.767 for MR24-DD, MR23-ED, and MR23-D, respectively.

A clear trend can be observed: models capable of processing longer context tend to achieve higher performance. This suggests that access to a broader portion of the user’s message history helps capture the evolution of linguistic signals more effectively, which is especially beneficial in user-level classification scenarios. RoBERTa-based models, despite being constrained by a shorter context window, still achieve competitive results.

Overall, the results demonstrate that the domain-adapted Spanish models clearly outperform their general-purpose models in all three user-level classification tasks. This confirms that incorporating mental-health-related data during pre-training leads to more effective representations for this domain. Models based on the Longformer architecture particularly benefit from their ability to process longer contexts, obtaining the best results across the three benchmarks. To the best of our knowledge, these models represent the first publicly available Spanish language models specifically designed for the mental health context.

However, these experiments evaluate the models under a conventional full-context classification setup, where the complete user history is available from the beginning. Early-detection tasks follow a very different evaluation protocol, requiring the model to make decisions incrementally as new messages arrive. In the following subsection, we analyze how the foundational models behave under this early-detection setting and whether their strong user-level performance translates to this more challenging scenario.

\subsection{Evaluation on Early Detection Tasks}

After assessing the models under a conventional full-context classification setup, we also evaluated their performance in early detection scenarios, where predictions must be issued incrementally as new user messages become available. This evaluation follows the official early-detection protocol described in Section~\ref{format_competi}, which imposes stricter constraints and reflects more realistic monitoring conditions. 

\subsubsection{Validation Results on Early-Detection Tasks}
\label{validation_FM}
We first evaluated the foundational models under the early-detection protocol, where predictions must be issued from the very first message. This setting reflects the natural, unconstrained behavior of the models. The results revealed apparent weaknesses: although the systems often trigger extremely early predictions, their accuracy is noticeably lower, and their decisions are volatile across tasks. This indicates that straightforward fine-tuning does not provide the models with any mechanism to regulate when a prediction should be made in an incremental setting.

To better understand this limitation, we carried out an additional study in which the models were required to wait for a predefined amount of text before issuing a decision. Specifically, we evaluated a series of minimum context thresholds (50, 100, 150, and 200 words), allowing us to analyze how performance evolves as progressively more initial evidence becomes available. This analysis enables the characterization of the early-detection behavior of the models and the determination of which starting points lead to more reliable decisions.

Tables~\ref{tab:windows_DD_Ava}, \ref{tab:windows_ED_Ava}, and \ref{tab:windows_D_Ava} report the complete validation results for all models, tasks, and minimum-context thresholds.

\begin{table}[htbp]
\centering
\footnotesize 
\renewcommand{\arraystretch}{1.45}
\setlength{\tabcolsep}{0.4em}
\caption{Performance on Validation partition of fine-tuned foundational models on MR24-DD across different minimum-context configurations (in words).}
\label{tab:windows_DD_Ava}
\begin{tabular}{l|c|cccc}
\hline
Model & Config & ERDE5 & ERDE30 & LTP & F1-score \\ \hline

RoBERTa-es-m-large & \nothreshold & 0.173 & 0.086 & \textbf{1} & 0.744 \\
                   & 50             & 0.213 & 0.063 & 4          & 0.805 \\
                   & 100            & 0.346 & 0.059 & 7          & 0.828 \\
                   & 150            & 0.422 & 0.064 & 10         & 0.844 \\
                   & 200            & 0.459 & 0.069 & 13         & 0.852 \\ \hline
Longformer-es-m-base & \nothreshold & 0.177 & 0.090 & \textbf{1} & 0.708 \\
                     & 50             & 0.193 & 0.048 & 4          & 0.856 \\
                     & 100            & 0.324 & 0.041 & 8          & 0.887 \\
                     & 150            & 0.414 & 0.049 & 11         & 0.893 \\
                     & 200            & 0.454 & 0.053 & 14         & 0.903 \\\hline
Longformer-es-m-large & \nothreshold & \textbf{0.171} & 0.073 & \textbf{1} & 0.776 \\
                      & 50             & 0.208 & 0.046 & 4          & 0.872 \\
                      & 100            & 0.334 & \textbf{0.040} & 8      & 0.873 \\
                      & 150            & 0.421 & 0.045 & 11         & 0.888 \\
                      & 200            & 0.450 & 0.049 & 14         & \textbf{0.904} \\ \hline

\end{tabular}
\end{table}

\begin{table}[htbp]
\centering
\footnotesize 
\renewcommand{\arraystretch}{1.45}
\setlength{\tabcolsep}{0.4em}
\caption{Performance on Validation partition of fine-tuned foundational models on MR23-ED across different minimum-context configurations (in words).}
\label{tab:windows_ED_Ava}
\begin{tabular}{l|c|cccc}
\hline
Model & Config & ERDE5 & ERDE30 & LTP & F1-score \\ \hline

RoBERTa-es-m-large & \nothreshold & 0.167 & 0.120 & \textbf{1} & 0.731 \\
                   & 50             & 0.258 & 0.098 & 6          & 0.801 \\
                   & 100            & 0.425 & 0.076 & 10         & 0.814 \\
                   & 150            & 0.467 & 0.081 & 13         & 0.814 \\
                   & 200            & 0.471 & 0.084 & 14         & 0.832 \\ \hline
Longformer-es-m-base & \nothreshold & 0.178 & 0.129 & \textbf{1} & 0.721 \\
                     & 50             & 0.228 & 0.094 & 5          & 0.798 \\
                     & 100            & 0.413 & 0.065 & 8          & 0.860 \\
                     & 150            & 0.456 & 0.070 & 11         & 0.870 \\
                     & 200            & 0.467 & 0.075 & 13         & 0.870 \\\hline
Longformer-es-m-large & \nothreshold & \textbf{0.157} & 0.127 & \textbf{1} & 0.731 \\
                      & 50             & 0.172 & 0.071 & 5          & 0.813 \\
                      & 100            & 0.365 & \textbf{0.041} & 8      & 0.932 \\
                      & 150            & 0.421 & 0.049 & 11         & \textbf{0.954} \\
                      & 200            & 0.441 & 0.053 & 13         & \textbf{0.954} \\ \hline

\end{tabular}
\end{table}

\begin{table}[htbp]
\centering
\footnotesize 
\renewcommand{\arraystretch}{1.45}
\setlength{\tabcolsep}{0.4em}
\caption{Performance on Validation partition of fine-tuned foundational models on MR23-D across different minimum-context configurations (in words).}
\label{tab:windows_D_Ava}
\begin{tabular}{l|c|cccc}
\hline
Model & Config & ERDE5 & ERDE30 & LTP & F1-score \\ \hline

RoBERTa-es-m-large & \nothreshold & \textbf{0.207} & 0.184 & \textbf{1} & 0.669 \\
                   & 50             & 0.272 & 0.169 & 5          & 0.714 \\
                   & 100            & 0.454 & \textbf{0.151} & 9      & 0.731 \\
                   & 150            & 0.500 & 0.159 & 12         & 0.737 \\
                   & 200            & 0.515 & 0.165 & 15         & 0.739 \\ \hline

Longformer-es-m-base & \nothreshold & 0.225 & 0.192 & \textbf{1} & 0.636 \\
                     & 50             & 0.301 & 0.179 & 5          & 0.678 \\
                     & 100            & 0.454 & 0.166 & 9          & 0.699 \\
                     & 150            & 0.474 & 0.173 & 12         & 0.712 \\
                     & 200            & 0.489 & 0.178 & 15         & 0.713 \\ \hline

Longformer-es-m-large & \nothreshold & 0.225 & 0.188 & \textbf{1} & 0.602 \\
                      & 50             & 0.334 & 0.181 & 6          & 0.637 \\
                      & 100            & 0.461 & 0.176 & 9          & 0.671 \\
                      & 150            & 0.489 & 0.176 & 12         & \textbf{0.740} \\
                      & 200            & 0.497 & 0.177 & 15         & \textbf{0.740} \\ \hline

\end{tabular}
\end{table}

Across all three tasks, the \nothresholdfirst configuration (when no minimum context is set) revealed apparent limitations of the foundational models. Because predictions must be made from the very first message, the systems typically triggered their first correct decision in the initial round (\textit{LTP} = 1). Although this led to favorable ERDE5 values, overall performance remained modest: F1-scores were noticeably lower, and the transition from \nothreshold to 50 words produced abrupt changes in all metrics. These results show that the models are susceptible to the amount of initial information, and that their behavior under minimal context is unstable and difficult to control.

When additional initial context is required, performance improved consistently. All models showed higher F1-scores and more stable ERDE30 values as the minimum-context threshold increased. For example, in MR24-DD, RoBERTa-es-m-large increased from F1 = 0.744 at \nothreshold to 0.828 at 100 words, and Longformer-es-m-base improved from 0.708 to 0.887. Similar trends appeared in MR23-ED and MR23-D. These improvements indicated that the models benefit substantially from having more information available at the start of the detection process. 

The 100-word threshold emerges as the most balanced configuration across tasks, primarily reflected in its ERDE30 behavior. In all three tasks, this setting consistently achieved the lowest ERDE30 values, indicating that it provided sufficient context to support reliable decisions without incurring excessive delay. F1-scores also reached their highest or near-highest values at this point, but with considerably smaller increments than those observed when moving from \nothreshold to 50 words. Beyond 100 words, additional context (150 or 200 words) produced only marginal improvements in F1 and generally led to slightly worse ERDE30 values, as predictions were postponed unnecessarily.

\label{threshold}
These validation results identify two configurations that best characterize the behavior of the foundational models. The \nothreshold setting reflects their natural, unconstrained operation and exposes the difficulties they face when working with minimal information. In contrast, we find the best balance between early detection and accuracy, reflected by the ERDE30 metric, at the threshold of 100 words. For this reason, we select both configurations throughout the remainder of the experimentation.

\subsubsection{Test Results on Early-Detection Tasks}

To assess the performance and generalization capabilities of the models under realistic early-detection conditions, we compared their results with the selected reference systems for each task.

Tables~\ref{tab:MR24_results_EAF}, \ref{tab:ED_results_EAF}, and \ref{tab:DEP_results_EAF} present the test results for all fine-tuned foundational models under these two configurations, together with the reference systems of each shared task.

\begin{table}[!ht]
\centering
\footnotesize 
\caption{MentalRisk24 - Disorder Detection (MR24-DD) Test results.}
\setlength{\tabcolsep}{2mm}
\renewcommand{\arraystretch}{1.45}
\begin{tabular}{lcccc}
\hline
                    Models   & \multicolumn{1}{l}{ERDE5} & \multicolumn{1}{l}{ERDE30} & \multicolumn{1}{l}{LTP} & \multicolumn{1}{l}{F1-score} \\ \hline
ELiRF-UPV                & 0,405                     & 0,045                      & 8                             & 0,874                         \\
UNED                     & \textbf{0,138}                     & 0,065                      & 2                             & 0,785                         \\
RoBERTa-base             & 0,162                     & 0,042                      & 3                             & 0,834                         \\ \hline
Roberta-es-m-large-\nothresholdp     & 0,193                     & 0,088                      & \textbf{1}                             & 0,740 \\
Longformer-es-m-base-\nothresholdp   & 0,163                     & 0,093                      & \textbf{1}                             & 0,705        \\
Longformer-es-m-large-\nothresholdp  & 0,169                     & 0,078                      & \textbf{1}                             & 0,772                \\ \hline
Roberta-es-m-large-100     & 0,342                     & 0,061                      & 7                             & 0,822 \\
Longformer-es-m-base-100   & 0,331                     & 0,044                      & 8                             & \textbf{0,880}       \\ 
Longformer-es-m-large-100  & 0,340                     & \textbf{0,041}                      & 8                             & 0,867                \\\hline
\end{tabular}
\label{tab:MR24_results_EAF}
\end{table}

In MR24-DD, presented in \cref{tab:MR24_results_EAF}, the 100-word configuration produced the highest results, with Longformer-es-m-base reaching an F1-score of 0.880 and outperforming all reference systems. ERDE30 values were also among the lowest, indicating that the models can provide accurate decisions with only moderate delay. The \nothreshold setting, however, remained clearly inferior across all metrics.

\begin{table}[!ht]
\centering
\footnotesize 
\caption{MentalRisk23 - Eating Disorders (MR23-ED) Test results.}
\setlength{\tabcolsep}{2mm}
\renewcommand{\arraystretch}{1.45}
\begin{tabular}{lcccc}
\hline
                    Models   & \multicolumn{1}{l}{ERDE5} & \multicolumn{1}{l}{ERDE30} & \multicolumn{1}{l}{LTP} & \multicolumn{1}{l}{F1-score} \\ \hline
CIMAT-NLP-GTO             & 0,334                     & \textbf{0,018}                      & 6                             & \textbf{0,966}                         \\
RoBERTa-large                  & \textbf{0,163}                     & 0,099                      & 2                             & 0,813                         \\
UNSL          & 0,433                     & 0,045                      & 8                             & 0,913                         \\ \hline
Roberta-es-m-large-\nothresholdp     & \textbf{0,163}                     & 0,124                      & \textbf{1}                             & 0,726 \\
Longformer-es-m-base-\nothresholdp   & 0,179                     & 0,137                      & \textbf{1}                             & 0,686            \\
Longformer-es-m-large-\nothresholdp  & 0,177                     & 0,149                      & \textbf{1}                             & 0,726                \\ \hline
Roberta-es-m-large-100     & 0,386                     & 0,081                      & 8                             & 0,826 \\
Longformer-es-m-base-100   & 0,386                     & 0,074                      & 9                            & 0,833            \\
Longformer-es-m-large-100  & 0,347                     & 0,049                      & 8                             & 0,920                \\ \hline
\end{tabular}
\label{tab:ED_results_EAF}
\end{table}

For the ED task, presented in the \Cref{tab:ED_results_EAF}, the best-performing system overall remained the CIMAT-NLP-GTO system, based on a classical TF-IDF + Naive Bayes model. Among transformers, the UNSL system established a strong reference system. At 100 words, Longformer-es-m-large achieved a F1-score of 0.920, surpassing all transformer reference systems, and an ERDE30 of 0.049, which is competitive though slightly above the UNSL score. This shows that, with sufficient initial context, the models were capable of reaching state-of-the-art performance among transformer architectures.

\begin{table}[!ht]
\centering
\footnotesize 
\caption{MentalRisk23 - Depression (MR23-D) Test results.}
\setlength{\tabcolsep}{2mm}
\renewcommand{\arraystretch}{1.45}
\begin{tabular}{lcccc}
\hline
                   Models   & \multicolumn{1}{l}{ERDE5} & \multicolumn{1}{l}{ERDE30} & \multicolumn{1}{l}{LTP} & \multicolumn{1}{l}{F1-score} \\ \hline
UMUTeam             & 0,548                     & 0,358                      & 30                             & \textbf{0,737}                         \\
VICOM-nlp                  & 0,275                     & 0,173                      & 2                             & 0,631                         \\
SINAI-SELA          & 0,395                     & \textbf{0,140}                      & 4                             & 0,720                        \\ \hline
Roberta-es-m-large-\nothresholdp     & 0,266                     & 0,236                      & \textbf{1}                             & 0,666 \\
Longformer-es-m-base-\nothresholdp   & 0,260                     & 0,209                      & \textbf{1}                             & 0,633                        \\
Longformer-es-m-large-\nothresholdp  & \textbf{0,250}                     & 0,198                      & \textbf{1}                             & 0,600                \\ \hline
Roberta-es-m-large-100     & 0,552                     & 0,223                      & 9                             & 0,726 \\
Longformer-es-m-base-100   & 0,467                     & 0,169                      & 10                             & 0,693  \\
Longformer-es-m-large-100  & 0,481                     & 0,186                      & 10                             & 0,666                \\ \hline
\end{tabular}
\label{tab:DEP_results_EAF}
\end{table}

The MR23-D, as shown in \Cref{tab:DEP_results_EAF}, demonstrated that the models exhibited more modest improvements. Still, the 100-word configuration yielded consistent gains, with RoBERTa-es-m-large reaching 0.726 F1 and reducing ERDE30 compared to the \nothreshold condition. The performance gap narrowed substantially once additional context was provided.

Across all tasks, the comparison with the reference systems revealed a consistent pattern. Under the \nothreshold configuration, the models tended to issue their first correct prediction in the initial round (\textit{LTP} = 1), which resulted in favorable ERDE5 values but noticeably lower F1 and ERDE30 performance. Their behavior under minimal context was unstable and strongly dependent on slight variations in the available evidence. In contrast, the 100-word configuration yielded substantial gains in F1 and more stable ERDE30 values, confirming the trend already observed in the validation phase: when provided with a moderate amount of initial context, the models achieved more reliable decisions and became competitive with the best systems submitted to the shared tasks.

Overall, the test results reveal a consistent pattern: the models perform poorly when required to make predictions from the very first message, but become highly competitive and, in several cases, surpass the reference systems when provided with a moderate amount of initial context. However, this improvement largely depends on selecting the appropriate minimum-context threshold, which is also a fixed value that does not adapt to each specific user's message history. These limitations underscore the need for a specialized training methodology specifically designed for early detection scenarios.

\section{Scenario-Specific Training Approach for Early Detection}
\label{sec:ice_method}

Datasets used for identifying symptoms of mental illness from social‑media posts typically comprise chronologically ordered user texts associated with a single label for the user.  Assigning the label at the user level reduces the information available for early-detection scenarios, as we are unable to determine which posts exhibit signs of symptoms for certain mental illnesses and which do not. This limitation restricts the utility of the data for training models designed for early detection. To adapt user-level datasets for early-detection scenarios, we propose the Incremental Context Expansion (ICE) methodology. ICE is based on generating and relabeling multiple training instances extracted from each user's history. This enables the model to learn when the accumulated context is sufficient to issue a reliable prediction without waiting for the full history.

\subsection{The Incremental Context Expansion Methodology}

The ICE methodology addresses a key challenge in relabeling data for early detection tasks: automatically identifying the minimal user context required to predict the user-level label. 
To generate relabeled data, ICE creates a set of new samples for each original dataset sample. These new samples consist of incrementally expanding partial contexts, each with a corresponding label.
For users exhibiting no symptoms, regardless of context expansion, the absence of symptoms remains constant, and all generated samples retain their original label. The situation is more complex for users who are symptomatic. Shorter contexts may lack sufficient information to reliably assess symptom presence, while longer contexts increase the likelihood of identifying such symptoms. Crucially, the length of context needed to detect a symptomatic user's illness will vary from individual to individual.

For symptomatic users, we need a stable and transparent mechanism to detect the transition from non-symptomatic to symptomatic contexts. We assume that context messages preceding the transition point do not contain symptom indicators, while those following it do. This assumption facilitates the creation of a relabeled dataset designed to train models for early detection, enabling them to either predict illness earlier or wait for more contextual information.
 
\begin{figure}[h]
    \centering
    \caption{Overview of the ICE workflow. The process generates incremental message contexts and assigns labels based on the transition point inferred with the transition detector.}
    \vspace{10pt}
    \includegraphics[width=0.65\linewidth]{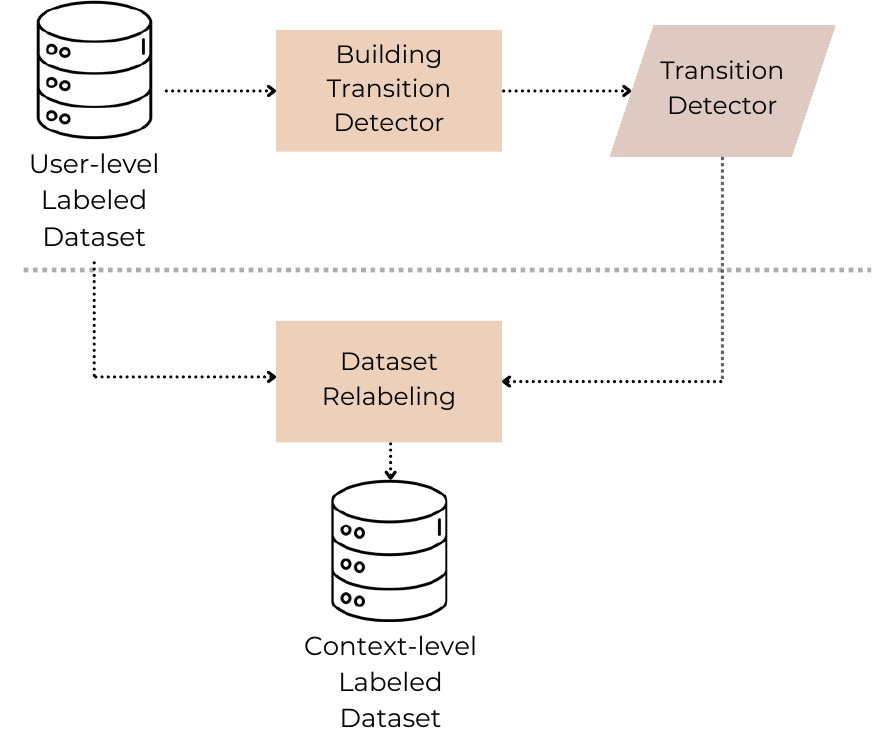}
    \label{fig:ice_pipeline}
\end{figure}

Figure \ref{fig:ice_pipeline} provides a visual overview of the ICE workflow, illustrating the two steps involved in this automatic relabeling process. In the first step, a transition detector is trained using the user-level labeled dataset. In the second step, the transition detector is used to relabel the user-level labeled dataset, obtaining a context-level labeled dataset.

For the first step, to identify the boundary between non-symptomatic and symptomatic states, we decided to use a SVM-based model due to its robustness in handling unlimited input length with a clear interpretability through decision boundaries, the detection of linguistic patterns, and its competitive results \cite{Casamayor2024}. This alignment enables the system to generalize well across varied user behaviors, maintaining both stability and transparency in its decision process. The SVM was trained as a user-level classifier with the original dataset. Later, the classifier was used to identify the transition point by classifying the incrementally expanding partial contexts from a user’s history until the prediction is changed from non-symptomatic to symptomatic.

\begin{figure}[h]
    \centering
    \caption{Generation of samples for symptomatic users in the ICE relabeling methodology. The red dashed line marks the transition point used to generate incremental samples.}
    \vspace{10pt}
    \includegraphics[width=0.75\linewidth]{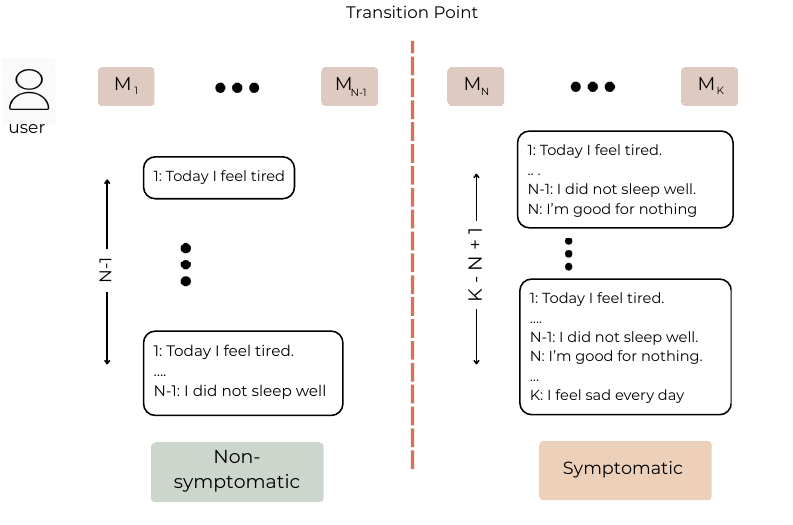}
    \label{fig:dataug}
\end{figure}

\cref{fig:dataug} illustrates the relabeling process for symptomatic users, as follows in step two. However, the general process follows these steps for any user-level labeled sample with $K$ messages:

\begin{enumerate}
    \item For symptomatic users, we use the transition detector to determine the transition point. If the transition point is not found by the detector, we assign the mean value of the transition points of all samples with a detected transition point in the dataset.

    \item For non-symptomatic users, we set the transition point to $K+1$.
    
    \item The $K$ context-level samples are generated as follows. Let $n$ represent the transition point, $m_i$ represents the ith message from the user, then:
    
    \begin{enumerate}
        \item We generate $(n - 1)$ non-symptomatic samples by gradually incrementing the context from the first message until the last message before the transition point:
        \begin{align*}
            m_1 \quad\quad \text{\scriptsize \textit{(sample 1)}}\\
            m_1 \oplus m_2 \quad\quad \text{\scriptsize \textit{(sample 2)}} \\
            \vdots \\
            m_1 \oplus ... \oplus m_{n-1} \quad\: \text{\scriptsize \textit{(sample n-1)}}
        \end{align*}
        \item and, if $n <= K$, then we generate $(K-n+1)$ symptomatic samples; from the transition point until the last message:
        \begin{align*}
            m_1 \oplus ... \oplus m_{n-1} \oplus m_{n}  \quad\: \text{\scriptsize \textit{(sample n)}}\\ 
            \vdots \\
            m_1 \oplus ...\oplus m_{n} \oplus ...\oplus m_{K-1} \oplus m_{K}  \quad\: \text{\scriptsize \textit{(sample K)}}
        \end{align*}
    \end{enumerate}
    
\end{enumerate}

\subsection{Training configuration}

\begin{figure}[H]
    \centering
    \caption{Fine-tuning workflow process that includes the ICE relabeling methodology.}
    \vspace{10pt}
    \includegraphics[width=0.65\linewidth]{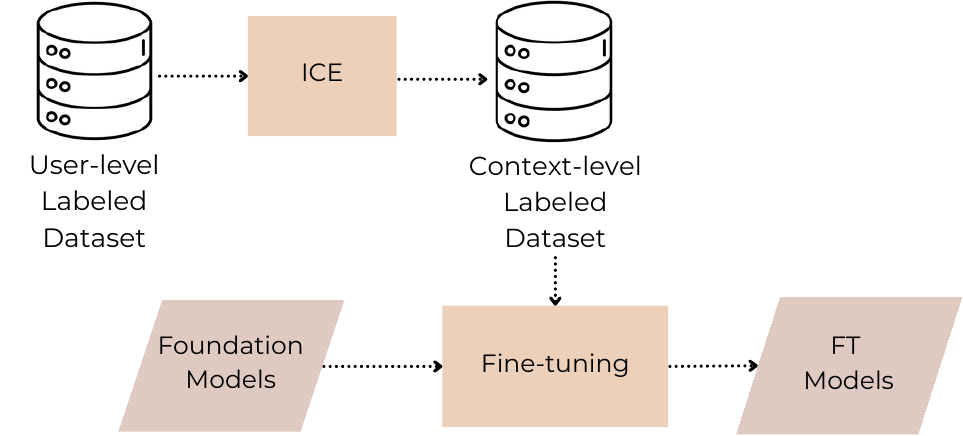}
    \label{fig:domain-specific-workflow}
\end{figure}

As we mentioned above, the ICE methodology produces a new dataset, context-level labeled dataset, adapted to the early detection scenario by relabeling the user-level dataset. This new dataset was integrated into a fine-tuning pipeline to obtain new mental health detection models tailored to early detection tasks. The complete workflow is shown in \Cref{fig:domain-specific-workflow}.

\begin{table}[!ht]
\footnotesize 
\centering
\caption{Amount of training samples of the user-level (original) and relabeled context-level datasets using ICE methodology.}
\setlength{\tabcolsep}{4mm}
\renewcommand{\arraystretch}{1.45}
\begin{tabular}{lccc}
\hline
\multicolumn{4}{c}{\textbf{MentalRisk24 - Disorder Detection (MR24-DD)}} \\ \hline
\textbf{Dataset} & None & Depression & Anxiety \\ \hline
user-level (original)     & 178  & 134  & 76  \\
context-level (ICE) & 7415 & 4108 & 2348 \\ \hline
\\
\hline
\multicolumn{4}{c}{\textbf{MentalRisk23 - Eating Disorders (MR23-ED)}} \\ 
\hline
\textbf{Dataset} & None & ED   \\ \hline
user-level (original)     & 85  & 63     \\
context-level (ICE)   & 2966 & 2049    \\ \hline
\\
\hline
\multicolumn{4}{c}{\textbf{MentalRisk23 - Depression (MR23-D)}} \\ \hline
\textbf{Dataset} & None & Depression  \\ \hline
user-level (original)     & 70  & 78     \\
context-level (ICE)   & 2920 & 2815    \\ \hline
\end{tabular}
\vspace{0.3cm}
\label{tab:datasets_samples}
\end{table}

The context-level labeled datasets increased the number of samples with respect to their corresponding user-level dataset, including both new positive and negative samples. \Cref{tab:datasets_samples} shows the comparison between the amount of training samples in the user-level and the context-level datasets for each task.  

For training each model, we used only those samples from the context-level dataset that fitted the input layer of the model. \Cref{tab:datasets_single} shows the amount of samples used for each task depending on each specific model architecture. Due to Longformer's ability to handle larger contexts, nearly all samples from the context-level labeled dataset were used.  

\begin{table}[!ht]
\footnotesize 
\centering
\caption{Amount of training samples of the context-level labeled datasets used for each model architecture.}
\setlength{\tabcolsep}{4mm}
\renewcommand{\arraystretch}{1.45}
\begin{tabular}{lccc}
\hline
\multicolumn{4}{c}{\textbf{MentalRisk24 - Disorder Detection (MR24-DD)}} \\ \hline
\textbf{Architecture} & None & Depression & Anxiety \\ \hline
RoBERTa & 5607 & 3000 & 1601 \\
Longformer   & 7415 & 4108 & 2348 \\ \hline
\\
\hline
\multicolumn{4}{c}{\textbf{MentalRisk23 - Eating Disorders (MR23-ED)}} \\ 
\hline
\textbf{Architecture} & None & ED   \\ \hline
RoBERTa & 2611 & 1629    \\
Longformer   & 2950 & 2008    \\ \hline
\\
\hline
\multicolumn{4}{c}{\textbf{MentalRisk23 - Depression (MR23-D)}} \\ \hline
\textbf{Architecture} & None & Depression  \\ \hline
RoBERTa & 1975 & 2060    \\
Longformer   & 2920 & 2815    \\ \hline
\end{tabular}
\vspace{0.3cm}
\label{tab:datasets_single}
\end{table}

To evaluate the impact of the context-level datasets on early detection tasks, we trained the same foundational models introduced in \cref{sec:foundational_models_m} using the same hyperparameter
shown in \cref{Table:Parametres_fine_tuning_1}. This way, a new set of classifiers specifically adapted to the incremental contexts provided by ICE was obtained (ICE models).

We intentionally retained this setup and the use of the same foundational models to ensure a consistent comparison between methodologies, avoiding improvements that could arise from hyperparameter tuning rather than from the relabeled dataset.

\subsection{Evaluation on Early Detection Tasks}

The ICE models were evaluated following the same early-detection framework described in Section~\ref{format_competi}. The protocol, metrics, reference systems, and minimum-context threshold configurations (Section \ref{threshold}) remain unchanged.

\subsubsection{Validation Results on Early-Detection Tasks}

The results of the ICE models on the validation sets, for the three MentalRisk tasks, are shown in  
Tables~\ref{tab:windows_ICE_DD_Ava}, \ref{tab:windows_ICE_ED_Ava}, and \ref{tab:windows_ICE_D_Ava}. Each table includes all metrics defined in the evaluation protocol, enabling a direct comparison across models and configurations.

\begin{table}[htbp]
\centering
\footnotesize 
\renewcommand{\arraystretch}{1.45}
\setlength{\tabcolsep}{0.4em}
\caption{Performance on Validation partition of ICE models on MR24-DD across different minimum-context configurations (in words).}
\label{tab:windows_ICE_DD_Ava}
\begin{tabular}{l|c|cccc}
\hline
Model & Config & ERDE5 & ERDE30 & LTP & F1-score \\ \hline

RoBERTa-es-m-large-ICE & \nothreshold & 0.125 & 0.052 & \textbf{4} & 0.888 \\
                       & 100            & 0.342 & 0.042 & 8          & 0.892 \\ \hline

Longformer-es-m-base-ICE & \nothreshold & 0.132 & 0.056 & \textbf{4} & 0.872 \\
                         & 100            & 0.341 & 0.037 & 9          & 0.898 \\ \hline

Longformer-es-m-large-ICE & \nothreshold & \textbf{0.121} & 0.049 & \textbf{4} & 0.876 \\
                          & 100            & 0.331 & \textbf{0.033} & 9      & \textbf{0.915} \\ \hline

\end{tabular}
\end{table}

\begin{table}[htbp]
\centering
\footnotesize 
\renewcommand{\arraystretch}{1.45}
\setlength{\tabcolsep}{0.4em}
\caption{Performance on Validation partition of ICE models on MR23-ED across different minimum-context configurations (in words).}
\label{tab:windows_ICE_ED_Ava}
\begin{tabular}{l|c|cccc}
\hline
Model & Config & ERDE5 & ERDE30 & LTP & F1-score \\ \hline

RoBERTa-es-m-large-ICE & \nothreshold & 0.121 & 0.053 & \textbf{4} & 0.900 \\
                       & 100            & 0.338 & 0.046 & 8          & 0.911 \\ \hline

Longformer-es-m-base-ICE & \nothreshold & \textbf{0.120} & 0.047 & \textbf{4} & 0.901 \\
                         & 100            & 0.339 & 0.039 & 9          & 0.921 \\ \hline

Longformer-es-m-large-ICE & \nothreshold & \textbf{0.120} & 0.048 & 5      & 0.902 \\
                          & 100            & 0.321 & \textbf{0.036} & 9  & \textbf{0.945} \\ \hline

\end{tabular}
\end{table}

\begin{table}[htbp]
\centering
\footnotesize 
\renewcommand{\arraystretch}{1.45}
\setlength{\tabcolsep}{0.4em}
\caption{Performance on Validation partition of ICE models on MR23-D across different minimum-context configurations (in words).}
\label{tab:windows_ICE_D_Ava}
\begin{tabular}{l|c|cccc}
\hline
Model & Config & ERDE5 & ERDE30 & LTP & F1-score \\ \hline

RoBERTa-es-m-large-ICE & \nothreshold & \textbf{0.198} & 0.159 & \textbf{3} & 0.715 \\
                       & 100            & 0.388 & \textbf{0.143} & 8      & \textbf{0.758} \\ \hline

Longformer-es-m-base-ICE & \nothreshold & 0.220 & 0.175 & \textbf{3} & 0.684 \\
                         & 100            & 0.421 & 0.161 & 9          & 0.732 \\ \hline

Longformer-es-m-large-ICE & \nothreshold & 0.201 & 0.165 & 4      & 0.699 \\
                          & 100            & 0.448 & 0.145 & 9      & \textbf{0.758} \\ \hline

\end{tabular}
\end{table}

Comparing these results with those shown in Tables~\ref{tab:windows_DD_Ava}, \ref{tab:windows_ED_Ava}, and \ref{tab:windows_D_Ava}, across the three tasks, the validation results of the ICE models revealed a consistent and more cautious behavior under early-detection conditions. In the \nothreshold configuration, all models delayed their first correct prediction to later rounds (with \textit{LTP} between 3 and 5), indicating that ICE effectively discourages premature decisions even when no explicit minimum context is imposed. This delayed triggering allowed the models to accumulate more informative evidence before committing, which was reflected in consistent gains across all metrics. These gains even surpassed the original results obtained with the 100-word configuration in most cases.

Generally speaking, models using the 100-word configuration achieved the best results in terms of ERDE30 and F1-score, while the best results in terms of strict early-detection (ERDE5 and LTP) were achieved by models using the \nothreshold configuration.

\subsubsection{Test Results on Early-Detection Tasks}

To evaluate the performance and generalization capabilities of the ICE models under realistic early-detection conditions, we compared their results with those of the selected reference systems for each task. Tables~\ref{tab:MR24_results_ICE}, \ref{tab:MR23_results_ED_ICE}, and \ref{tab:MR23_results_D_ICE} report the performance of the three reference systems and the ICE models with \nothreshold and 100-word configurations across the three tasks.

\begin{table}[!ht]
\footnotesize 
\centering
\caption{MentalRisk24 - ICE models Test results on Disorder Detection (MR24-DD) task}
\setlength{\tabcolsep}{0.5em}
\renewcommand{\arraystretch}{1.45}
\begin{tabular}{lcccc}
\hline
                    Models   & \multicolumn{1}{l}{ERDE5} & \multicolumn{1}{l}{ERDE30} & \multicolumn{1}{l}{LTP} & \multicolumn{1}{l}{F1-score} \\ \hline
ELiRF-UPV                & 0,405                     & 0,045                      & 8                             & 0,874                         \\
UNED                     & 0,138                     & 0,065                      & \textbf{2}                             & 0,785                         \\
RoBERTa-base             & 0,162                     & 0,042                      & 3                             & 0,834                         \\ \hline
RoBERTa-es-m-large-ICE-\nothresholdp     & 0,135                     & 0,051                      & 4                             & 0,885 \\
Longformer-es-m-base-ICE-\nothresholdp   & 0,135                     & 0,055                      & 4                             & 0,868                         \\
Longformer-es-m-large-ICE-\nothresholdp  & \textbf{0,133}                     & 0,047                      & 4                             & 0,885                \\
\hline
RoBERTa-es-m-large-ICE-100     & 0,345                     & 0,056                      & 8   & 0,885 \\
Longformer-es-m-base-ICE-100   & 0,331                     & 0,043                      & 8                             & 0,892                         \\
Longformer-es-m-large-ICE-100  & 0,340                     & \textbf{0,040}                      & 9                             & \textbf{0,905}                \\ 
\hline
\end{tabular}
\label{tab:MR24_results_ICE}
\end{table}

Table~\ref{tab:MR24_results_ICE} reports well-balanced performance across all metrics for the ICE models on the MR24-DD test set. Notably, under the \nothreshold configuration, the three architectures demonstrated competitive ERDE5 values and F1-scores, even improving, in most cases, the reference systems. This confirms that the behavioral patterns observed during validation generalize to new data, and the ICE models learn to delay symptomatic predictions until sufficient evidence is available. The 100-word configurations further improve ERDE30 and F1-score values, but decrease the early detection capabilities of the models (ERDE5 and LTP metrics) compared to \nothreshold configurations.

\begin{table}[!ht]
\centering
\caption{MentalRisk23 - Eating Disorder  (MR23-ED) ICE results.}
\setlength{\tabcolsep}{0.5em}
\footnotesize 
\renewcommand{\arraystretch}{1.45}
\begin{tabular}{lcccc}
\hline
                    Models   & \multicolumn{1}{l}{ERDE5} & \multicolumn{1}{l}{ERDE30} & \multicolumn{1}{l}{LTP} & \multicolumn{1}{l}{F1-score} \\ \hline
CIMAT-NLP-GTO             & 0,334                     & \textbf{0,018}                      & 6                             & \textbf{0,966}                         \\
RoBERTa-large                  & 0,163                     & 0,099                      & 5                             & 0,813                         \\
UNSL          & 0,433                     & 0,045                      & 8                             & 0,913                         \\ \hline
RoBERTa-es-m-large-ICE-\nothresholdp     & 0,134                     & 0,060                      & \textbf{4}                             & 0,893 \\
Longformer-es-m-base-ICE-\nothresholdp   & \textbf{0,133}                     & 0,064                      & \textbf{4}                             & 0,880                         \\
Longformer-es-m-large-ICE-\nothresholdp  & \textbf{0,133}                     & 0,059                      & \textbf{4}                             & 0,886                \\
\hline
RoBERTa-es-m-large-ICE-100     & 0,355                     & 0,056                      & 8   & 0,893 \\
Longformer-es-m-base-ICE-100   & 0,361                     & 0,052                      & 9                             & 0,886                         \\
Longformer-es-m-large-ICE-100  & 0,343                     & 0,043                      & 9                             & 0,923                \\ \hline
\end{tabular}
\label{tab:MR23_results_ED_ICE}
\end{table}

\Cref{tab:MR23_results_ED_ICE} shows the results for task MR23-ED. The test results confirm the generalization of the results obtained in validation, showing better ERDE30 and F1 results for the 100‑word configurations and better ERDE5 and LTP results for the \nothreshold configurations. The ICE models outperformed reference models except CIMAT‑NLP‑GTO, a classic machine‑learning approach that uses a decision threshold to assign the positive class. This threshold, tuned specifically for the task, causes the model to wait until it has sufficient context (greater certainty) before making a prediction. As a result, the model achieves superior performance on detection metrics such as ERDE30 and F1‑score, but it does not excel in strict early‑detection metrics (ERDE5 and LTP) where the ICE models with the \nothreshold configuration perform best.

\begin{table}[!ht]
\centering
\caption{MentalRisk23 - Depresion (MR23-D) ICE results.}
\setlength{\tabcolsep}{0.5em}
\footnotesize 
\renewcommand{\arraystretch}{1.45}
\begin{tabular}{lcccc}
\hline
                    Models   & \multicolumn{1}{l}{ERDE5} & \multicolumn{1}{l}{ERDE30} & \multicolumn{1}{l}{LTP} & \multicolumn{1}{l}{F1-score} \\ \hline
UMUTeam             & 0,548                     & 0,358                      & 30                             & 0,737                         \\
VICOM-nlp                  & 0,275                     & 0,173                      & \textbf{2}                             & 0,631                         \\
SINAI-SELA          & 0,395                     & 0,140                      & 4                             & 0,720                        \\ \hline
RoBERTa-es-m-large-ICE-\nothresholdp     & \textbf{0,225}                     & 0,188                      & 3                             & 0,696 \\
Longformer-es-m-base-ICE-\nothresholdp   & 0,246                     & 0,189                      & 3                             & 0,663                         \\
Longformer-es-m-large-ICE-\nothresholdp  & 0,242                     & 0,192                      & 3                             & 0,646                \\
\hline
RoBERTa-es-m-large-ICE-100     & 0,426                     & \textbf{0,139}                      & 8   & \textbf{0,753} \\
Longformer-es-m-base-ICE-100   & 0,436                     & 0,161                      & 9                             & 0,726                         \\
Longformer-es-m-large-ICE-100  & 0,458                     & 0,163                      & 9                             & 0,696                \\ \hline
\end{tabular}
\label{tab:MR23_results_D_ICE}
\end{table}

The MR23‑D task proves to be the most challenging of the three tasks, as shown in \Cref{tab:MR23_results_D_ICE}. All systems exhibited lower overall performance, likely due to the subtle, indirect, and context-dependent nature of depression-related language that seems to be harder to capture with automatic methods. Despite this complexity, ICE models remained competitive in terms of performance, and the best ICE models outperformed the reference models. Furthermore, both minimum-context configurations (\nothreshold and 100-word) altered the performance behavior in the same way as in the previously analyzed task.

Across the three MentalRisk tasks, the ICE models show consistent performance under early-detection conditions. In the \nothreshold configuration, the models achieve competitive F1-scores while obtaining lower ERDE5 values and lower LTP than the reference systems, indicating better and earlier decisions. When evaluated under the 100-word configuration, all models improve their ERDE30 and F1-score values, reaching their best overall performance across tasks. Notably, \nothreshold configuration shows better performance for strict early detection metrics (ERDE5 and LTP), retaining competitive performance in detection metrics (ERDE30 and F1-score), without the need to set a specific-task minimum amount of context.

\section{Discussions}
\label{sec:discussions}

The results presented in the previous sections enable us to examine the behavior of the proposed models from a broader perspective and reflect on the impact of the ICE methodology on early detection performance. In this section, we analyze the overall tendencies observed throughout the study and discuss their implications.

\subsection{Overview of the Findings}

Across all experiments, several consistent patterns emerge. The models fine-tuned with the original datasets achieve strong results under full-context classification. Still, their behavior becomes unstable when evaluated under early-detection constraints: they tend to issue predictions too early, before sufficient evidence has accumulated. This limitation underscores the need for methodologies that not only focus on what to predict, but also on when to make predictions.

The ICE methodology directly addresses this issue by exposing the models to sequences of incremental expanding contexts during training. As a result, the fine-tuned models with ICE-generated data exhibit a more regulated decision process: they are better able to delay their symptomatic prediction when the available evidence is insufficient and to commit earlier when clear signals appear. 

Regarding the two context configurations, the \nothreshold configuration delivers superior early‑detection metrics and maintains good overall detection performance by relying on very limited, immediate signals. In contrast, the 100‑word configuration boosts overall detection performance by allowing a more delayed decision. In light of the results obtained in validation and testing, we observe that the ICE methodology improves the overall performance of the models, regardless of the minimum-context configuration chosen.

\subsection{The impact of ICE under different minimum-context configurations}

Based on the prior results, it has been established that imposing a minimal context forces the models to wait for more evidence, which in turn improves their performance. It was also established that the ICE methodology enabled the creation of incremental datasets, which boosted model accuracy. In this section, we will examine how the combination of the two factors, using or not using the ICE methodology and employing a minimal context, affects model performance for both classification and early‑detection metrics.

To better understand the practical effect of ICE on decision dynamics, we analyze the evolution of ERDE5, ERDE30, and F1-score metrics as a function of the minimum required context (0-200 words). \Cref{fig:ERDE5_TEST,fig:ERDE30_TEST,fig:F1_TEST} compare ICE and non-ICE models for each architecture on the test set across the three tasks, providing a direct view of how performance changes as more evidence becomes available.

\subsubsection{Strict early-detection performance (ERDE5)}

\begin{figure}[!ht]
    \centering
    \caption{ERDE5 evolution across the different methodologies and tasks in the test partition.}
    \label{fig:ERDE5_TEST}

    \begin{minipage}[t]{0.48\textwidth}
        \centering
        \includegraphics[width=\linewidth]
        {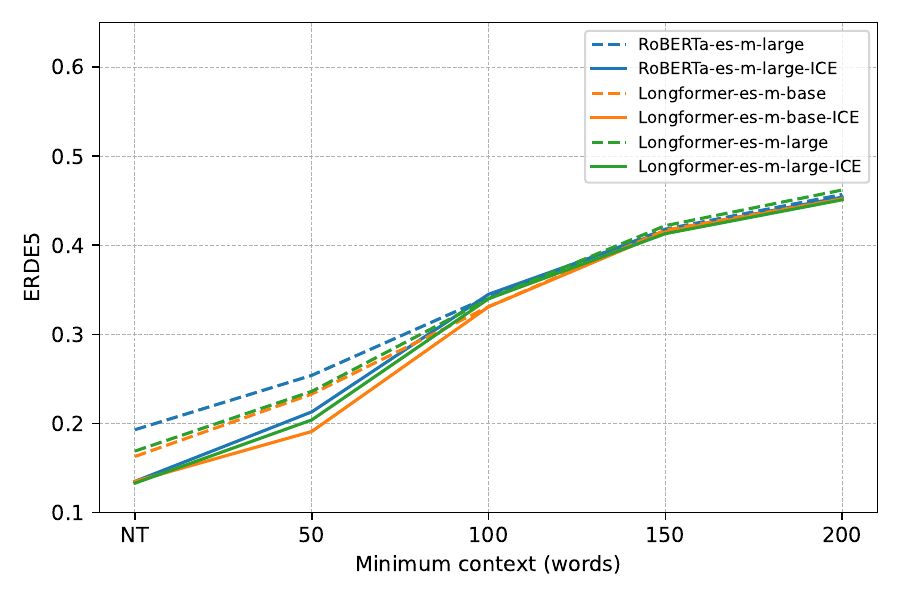}

        \footnotesize\textbf{(a)} MR24-DD
    \end{minipage}
    \hfill
    \begin{minipage}[t]{0.48\textwidth}
        \centering
        \includegraphics[width=\linewidth]
        {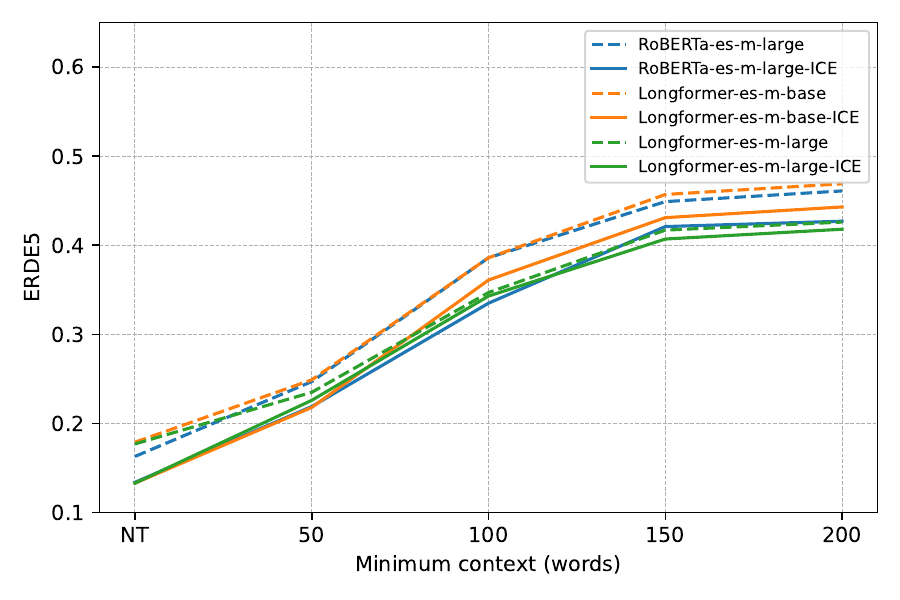}

        \footnotesize\textbf{(b)} MR23-ED
    \end{minipage}

    \vspace{4mm}

    \begin{minipage}[t]{0.48\textwidth}
        \centering
        \includegraphics[width=\linewidth]
        {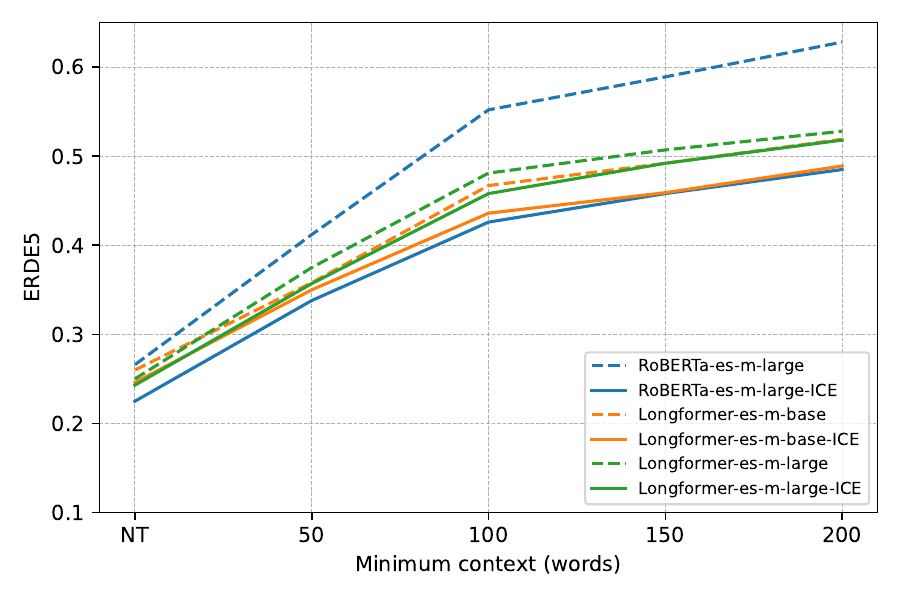}

        \footnotesize\textbf{(c)} MR23-D
    \end{minipage}
\end{figure}

The ERDE metric serves as a time efficiency evaluator, penalizing detection delays by integrating the number of posts required to trigger an alert as a critical factor in the error calculation. Within this framework, Figure~\ref{fig:ERDE5_TEST} shows that the performance of all the models in terms of ERDE5 degrades rapidly as the minimum required context increases. Notably, ICE models almost systematically outperform their non-ICE counterparts in most configurations and tasks.

The rapid deterioration of the ERDE5 metric, especially evident when the minimum context reaches the 100-word threshold, is a direct consequence of the interaction between the penalty function of the metric and the low information density of the datasets. In the three tasks evaluated, the average post length is significantly short, ranging between 14 and 15 tokens per post (Tables~\ref{tab:est_DD}, \ref{tab:est_ED}, and \ref{tab:est-D}). To meet a requirement of 100 words of context, the system is forced to process at least 8 or 9 consecutive posts. Since the ERDE5 metric sets its tolerance threshold at 5 posts, any decision based on context exceeding 50 words is penalized by the metric. Consequently, the marked increase in ERDE5 beyond 100 words of context reflects the exhaustion of the "latency budget" allowed by the metric. Once the accumulation of context exceeds the 5-post limit, for true positives, the ERDE5 rapidly increments the delay penalization. This behavior highlights that, in scenarios with sparse data and complex tasks, the margin for achieving a low ERDE5 score is extremely narrow.

According to this analysis, it is reasonable to conclude that the \nothreshold configuration consistently yields the lowest ERDE5 values. However, the performance of ICE models is noteworthy. As observed in Section~\ref{validation_FM}, non-ICE models tend to rush into classifying positives with LTP = 1. While this early classification increases false positives, the ERDE5 metric does not penalize them as severely as it does late true positives. Nevertheless, ICE models achieve the best results in \nothreshold configurations, thanks to their ability to make reliable early predictions by minimizing unnecessary waiting time without incurring the rush of the non-ICE models. By making most of their predictions within the critical window of the first 5 turns (LTP 3 or 4), ICE models get a better trade-off between prudent classification and the requirement for immediacy. 

It is also important to note that, while the results for MR24-DD and MR23-ED show a similar trend, performance on MR23-D is considerably lower. This discrepancy is consistent with the greater intrinsic difficulty of this task, as observed in previous results.

\subsubsection{Soft early-detection performance (ERDE30)}

\begin{figure}[!ht]
    \centering
    \caption{ERDE30 evolution across the different methodologies and tasks in the test partition.}
    \label{fig:ERDE30_TEST}

    \begin{minipage}[t]{0.48\textwidth}
        \centering
        \includegraphics[width=\linewidth]
        {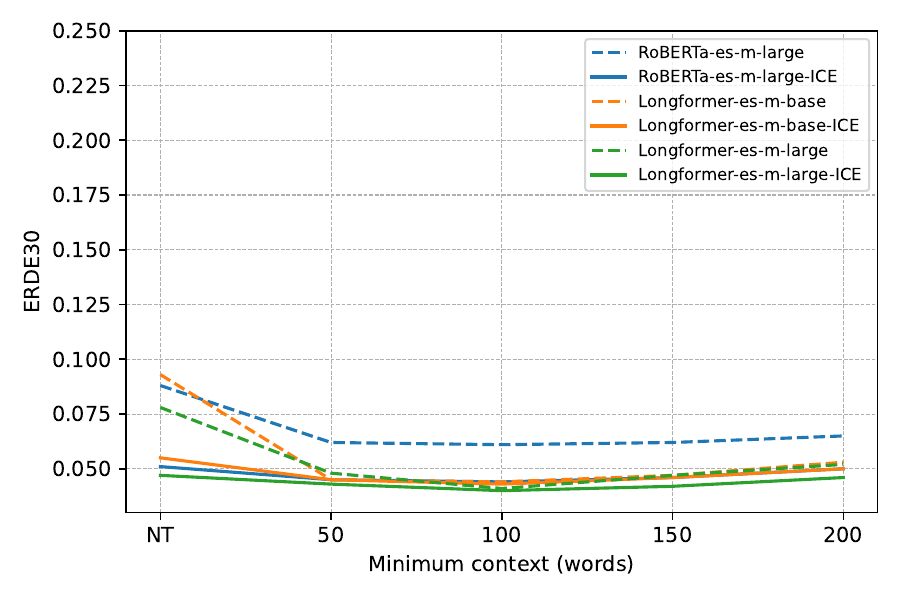}

        \footnotesize\textbf{(a)} MR24-DD
    \end{minipage}
    \hfill
    \begin{minipage}[t]{0.48\textwidth}
        \centering
        \includegraphics[width=\linewidth]
        {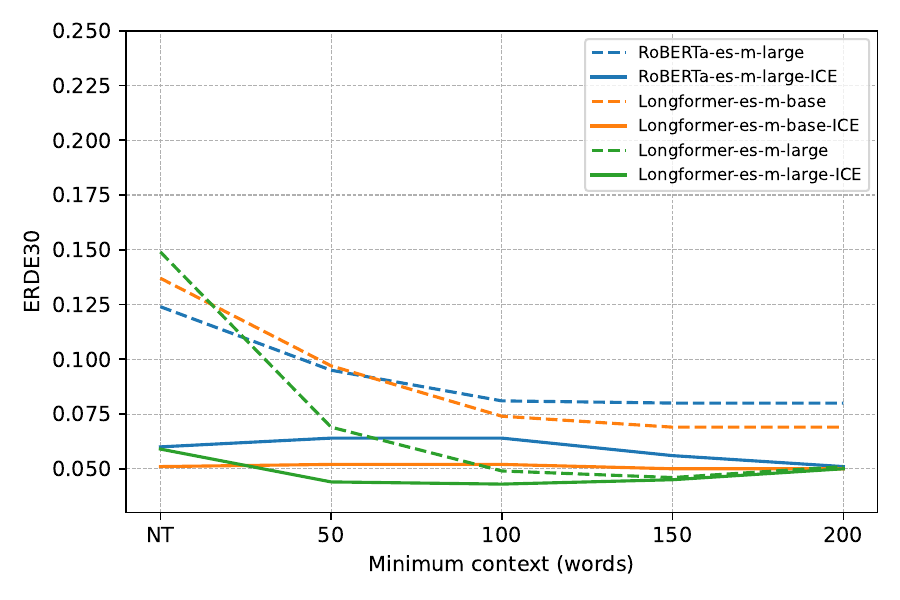}

        \footnotesize\textbf{(b)} MR23-ED
    \end{minipage}

    \vspace{4mm}

    \begin{minipage}[t]{0.48\textwidth}
        \centering
        \includegraphics[width=\linewidth]
        {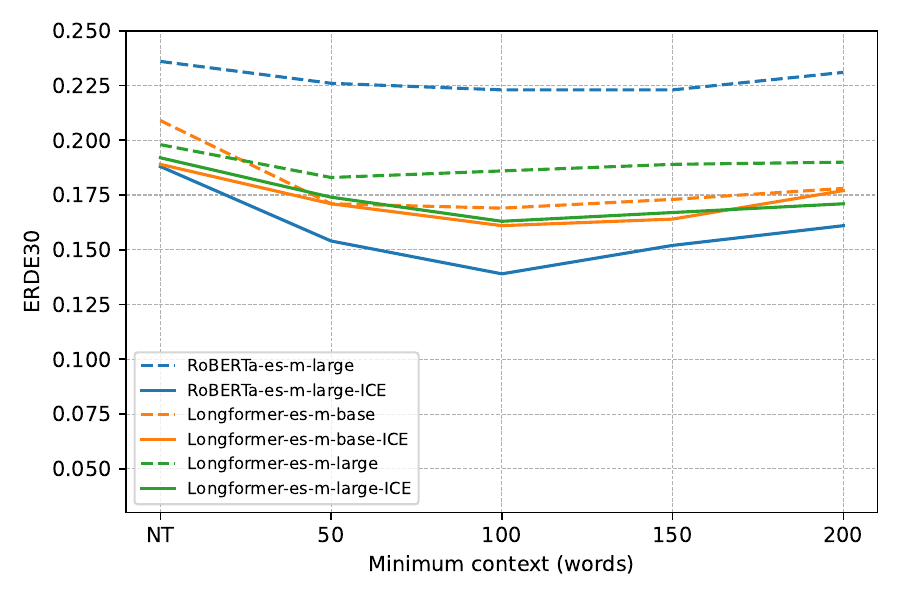}

        \footnotesize\textbf{(c)} MR23-D
    \end{minipage}
\end{figure}

Similar to ERDE5, the ERDE30 metric serves as a temporal efficiency evaluator that penalizes detection delay. However, it is important to note that ERDE30 is not the most representative indicator for evaluating early detection in these specific tasks, given that users have a significantly reduced average number of posts; the average is around 36 posts in MR24-DD and MR23-D, while in MR23-ED, the average drops to 31 posts (Tables~\ref{tab:est_DD}, \ref{tab:est_ED}, and \ref{tab:est-D}). Since the average number of posts is so close to the penalty threshold (30), the metric loses some of its ability to discriminate between actual precocity and near misses, as almost any decision based on the full history nears the temporal budget limit. In this framework, the ERDE30 metric is more closely related to the performance of the models as classifiers than to their performance as early detection evaluators.

As illustrated in Figure \ref{fig:ERDE30_TEST}, it is evident that non-ICE models show a more inconsistent behavior, starting with notably deficient ERDE30 values in the \nothreshold configuration. These models improve significantly when a 50-word context is introduced, and tend to stabilize at 100 words. Beyond this point, their performance typically degrades slightly towards 200 words, with the sole exception of the MR23-ED task, where such degradation does not occur.

The ICE models demonstrate a much more robust learning dynamic, represented by significantly lower and flatter curves. In MR24-DD, the models optimize their performance at the 50 and 100-word thresholds before worsening due to latency. For the MR23-ED task, the most excellent stability is observed, highlighting that the three ICE architectures converge to identical performance upon reaching 200 words.

Furthermore, the MR23-D task presents the poorest results of the study, with much higher error values compared to the other tasks, confirming its greater intrinsic difficulty. In this task, even the ICE models suffer a rapid and marked deterioration after exceeding the 100-word threshold.

\subsubsection{Overall detection performance (F1-score)}

\begin{figure}[!ht]
    \centering
    \caption{F1-score evolution across the different methodologies and tasks in the test partition.}
    \label{fig:F1_TEST}

    \begin{minipage}[t]{0.48\textwidth}
        \centering
        \includegraphics[width=\linewidth]
        {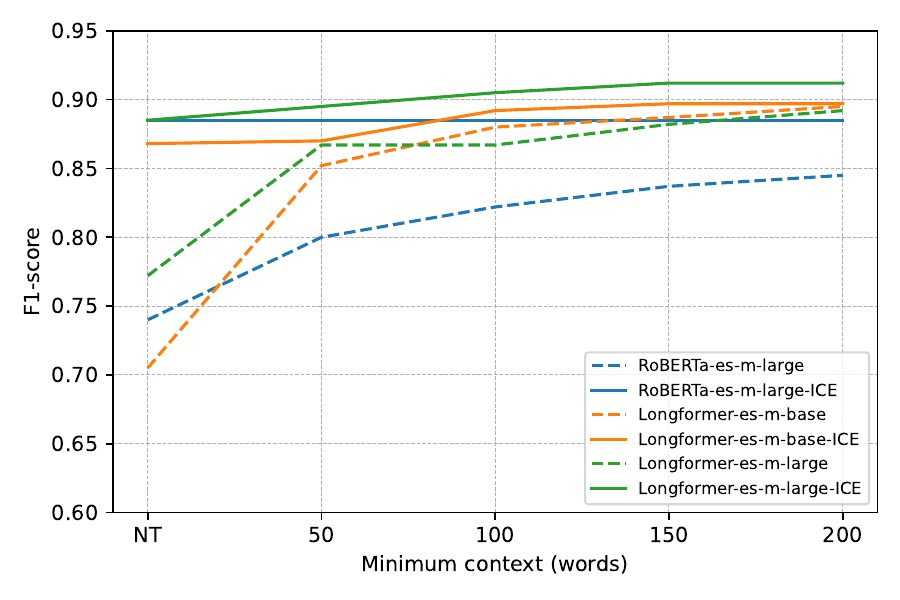}

        \footnotesize\textbf{(a)} MR24-DD
    \end{minipage}
    \hfill
    \begin{minipage}[t]{0.48\textwidth}
        \centering
        \includegraphics[width=\linewidth]
        {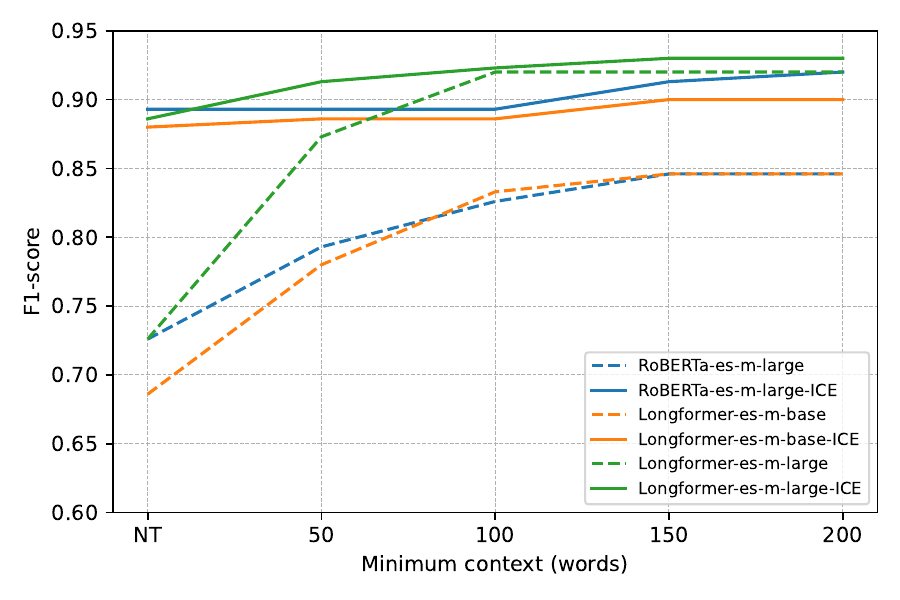}

        \footnotesize\textbf{(b)} MR23-ED
    \end{minipage}

    \vspace{4mm}

    \begin{minipage}[t]{0.48\textwidth}
        \centering
        \includegraphics[width=\linewidth]
        {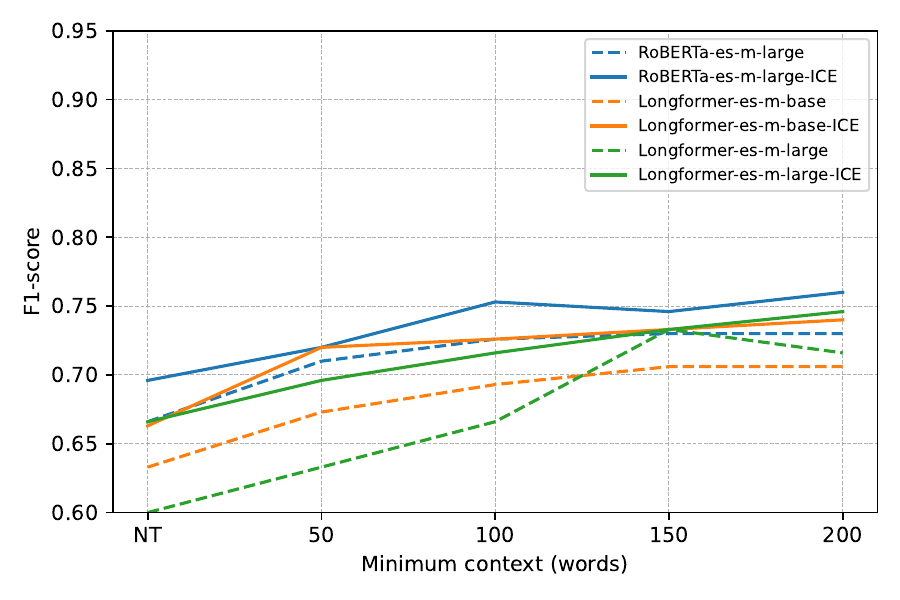}

        \footnotesize\textbf{(c)} MR23-D
    \end{minipage}
\end{figure}

As with previous metrics, the evolution of the F1-score in response to the increase in the minimum required context, Figure \ref{fig:F1_TEST}, provides clear evidence of the stabilization effect introduced by the ICE methodology. Since the F1-score metric evaluates the pure classification capability of the models, it is expected that, as the context increases, the models will have more information and, therefore, achieve better results. However, in this analysis, the absolute value of the F1-score at 200 words is not as crucial as its trajectory and stability throughout the process.

In the non-ICE models, the F1-score exhibits marked fluctuations in the initial stages, with abrupt performance jumps as soon as additional context becomes available. This behavior indicates that these models lack stability, as they base their predictions on insufficient information fragments. This means that any small change in the initial data drastically alters the outcome, leading to unreliable diagnoses.

In contrast, ICE models exhibit significantly more linear and smoother F1-score trajectories across all tasks. Their performance in the \nothreshold configuration is already highly competitive, and subsequent improvements are progressive, indicating a much more robust confidence. Only for the MR23-D task, the increase in minimum-context continues to boost the performance of the models. These results reflect a structural shift in decision strategy. By avoiding symptomatic samples during training until sufficient evidence has accumulated, ICE models learn to avoid errors from premature decisions. This results in more robust models.

It is worth noting that, in most cases, ICE models perform considerably better even at the 200-word configuration. This demonstrates that training using ICE not only optimizes early detection but also enhances the final discrimination capability of the models, enabling them to achieve a higher performance ceiling than conventionally trained models.

\subsection{The Impact of ICE Under the Different Architectures}

As stated above, the most significant impact of ICE is its ability to drastically improve model performance in low-context scenarios. Across all three metrics, the models utilizing ICE consistently outperform their non-ICE counterparts when no minimum context is set (NT configuration), regardless of their architecture. In Figure~\ref{fig:F1_TEST}, for instance, the F1-score for RoBERTa and Longformer models fine-tuned with ICE starts substantially higher than those without it.

Furthermore, ICE acts as a stabilizer that minimizes the performance gap between different model sizes and architectures. In the ERDE plots (Figures~\ref{fig:ERDE5_TEST} and \ref{fig:ERDE30_TEST}), RoBERTa-large and Longformer-base often converge or follow very similar trajectories when ICE is applied, whereas their non-ICE counterparts show much higher variance and error rates. This indicates that ICE helps smaller or less complex architectures achieve results comparable to larger models, specifically by reducing the penalty for early detection. By flattening the error curves and lifting the F1-score plateaus, ICE proves to be a critical enhancement for making these models more reliable and faster to respond in real-world risk assessment tasks.

\subsection{Limitations}

While the presented work has demonstrated competitive performance in early detection tasks, several limitations of this work should be acknowledged:

\begin{itemize}
    \item \textbf{Data bias and domain coverage.} The available Spanish mental health corpora remain limited in size and diversity, mostly derived from social media sources such as Reddit and Telegram. This introduces potential demographic and linguistic biases that may affect the generalization of findings to clinical or other real-world contexts.
    
    \item \textbf{Machine translation artifacts.} Although the use of automatically translated data (via EasyNMT) increased data availability to develop the foundational models, it may have introduced subtle semantic drift or loss of affective nuance. These effects could be particularly relevant in tasks such as depression detection, where small linguistic variations can alter the interpretation of emotional intent.

    \item \textbf{Dependence on transition detector in ICE methodology.} The ICE methodology relies on a transition detector, implemented in this work using an SVM classifier, to identify when the accumulated messages of a positive user begin to show evidence of pathology. If this criterion fails to capture early signals or assigns relevance inconsistently, the resulting noise propagates to the incremental training samples. This limitation may be particularly critical in tasks where initial symptoms are linguistically diffuse or implicit.
    
    \item \textbf{Interpretability and explainability.} This work focuses on quantitative evaluation and does not analyze the interpretability or transparency of the proposed models. Since early-detection systems can have implications for clinical decision-making or mental-health monitoring, understanding the rationale behind model predictions remains an essential aspect for future deployment.
\end{itemize}

\section{Future Work}
\label{sec:future_work}

Several promising research avenues remain open for exploration after this work. These directions directly address the main limitations identified in this study and aim to further enhance the interpretability and robustness of early detection systems.

\begin{itemize}
    \item \textbf{Data Augmentation:}
    We plan to complement the automatic relabeling method of ICE by including data augmentation techniques. We aim to create synthetic context or randomly modify the original context, thereby diversifying the training data and addressing the limitations of small datasets. This process will incorporate uncertainty estimation and ultimately improve model robustness in early detection tasks.

    \item \textbf{Explainable early detection with long contexts:}
    Understanding the reasoning behind model predictions is essential for deployment in sensitive domains. Integrating explainability tools, such as SHAP \cite{lundberg2017shap} or LIME \cite{ribeiro2016lime}, into long-sequence models could reveal which parts of a user’s history most influence each decision, thereby facilitating clinician trust and interpretability.

    \item \textbf{Improving the ICE methodology:}
    The current ICE methodology depends on the decision points of a transition detector to define symptom onset. Future work could explore end-to-end training where the detector and the base model are optimized together. Reinforcement learning could be used to reward accurate and early detections while penalizing false positives or excessive delays, fostering a dynamic balance between speed of detection and precision \cite{zhou2019rlnlp}. 

\end{itemize}

\section{Conclusions}
\label{sec:conclusions}

This work introduced three Spanish-language foundational models tailored to the mental health domain. Through domain-adaptive pre-training on an extended dataset of Spanish mental health texts, the models learned representations that capture the linguistic and affective patterns characteristic of this domain.

We have proposed the ICE automatic relabeling methodology to adapt user-level labeled datasets to early detection tasks by generating new context-level labeled datasets. ICE is based on generating and relabeling multiple training instances extracted from each user's history. To evaluate the impact of ICE on the performance of models fine-tuned with the generated context-level labeled dataset, we have conducted experiments on three benchmark early-detection tasks.

The results validate that ICE methodology allows models to dynamically approach the optimal operating point, drastically reducing error with few context maintaining a constant competitive advantage over non-ICE models, which depend critically on fixed context thresholds to avoid mediocre results.
In practice, ICE methodology allows for improved classification quality by waiting only the time strictly necessary to consolidate significant evidence, without introducing excessive latency. This behavior distinguishes a standard classifier from an effective early detection system, validating ICE as an methodology that enhances both initial reliability and overall system accuracy.

By combining the presented foundational models with the ICE methodology, we improved the state-of-the-art in most Spanish early-detection tasks addressed.

\bibliographystyle{apalike}
\bibliography{biblio} 

\section*{Acknowledgment}

This work was partially supported by the Spanish Agencia Estatal de
Investigación (AEI) under grant PID2024-155948OB-C55, within the
Proyectos de Generación de Conocimiento program, and co-funded by the
European Regional Development Fund (ERDF) under the Pluriregional
Operational Programme Spain FEDER 2021--2027.

This work was also partially supported by
MCIN/AEI/10.13039/501100011033 and by the European Regional Development
Fund, ``A way of making Europe,'' under grant PID2021-126061OB-C41.

Andreu Casamayor-Segarra was supported by the Spanish Ministerio de
Universidades under grant FPU24/01938. This work was also partially
supported by the Spanish Ministerio de Universidades under grant
FPU21/05288 for university teacher training and by the Generalitat
Valenciana under project CIPROM/2021/023.

\end{document}